# AI/ML Algorithms and Applications in VLSI Design and Technology

Deepthi **Amuru**[a,∗], Harsha V. **Vudumula**[a], Pavan K **Cherupally**[a], Sushanth R **Gurram**[a], Amir **Ahmad**[b], Andleeb **Zahra**[a] and Zia **Abbas**[a]

[a]*Center for VLSI and Embedded Systems Technology (CVEST), International Institute of Information Technology, Hyderabad (IIIT-H), Gachibowli, Hyderabad, 500032, India*
[b]*College of IT, UAE University, Al Ain, 15551, Abu Dhabi, UAE*

**ABSTRACT**

An evident challenge ahead for the integrated circuit (IC) industry is the investigation and development of methods to reduce the design complexity ensuing from growing process variations and curtail the turnaround time of chip manufacturing. Conventional methodologies employed for such tasks are largely manual, time-consuming, and resource-intensive. In contrast, the unique learning strategies of artificial intelligence (AI) provide numerous exciting automated approaches for handling complex and data-intensive tasks in very-large-scale integration (VLSI) design and testing. Employing AI and machine learning (ML) algorithms in VLSI design and manufacturing reduces the time and effort for understanding and processing the data within and across different abstraction levels. It, in turn, improves the IC yield and reduces the manufacturing turnaround time. This paper thoroughly reviews the AI/ML automated approaches introduced in the past toward VLSI design and manufacturing. Moreover, we discuss the future scope of AI/ML applications to revolutionize the field of VLSI design, aiming for high-speed, highly intelligent, and efficient implementations.

## 1. Introduction

A dramatic revolution has been triggered in the field of electronics by the advent of complementary metal-oxide-semiconductor (CMOS) transistors in the integrated circuit (IC) industry, leading to the era of semiconductor devices. Thenceforth, CMOS technology has been the predominant technology in the field of microelectronics. The number of transistors fabricated on a single chip has increased exponentially since the 1960s [1], [2]. The continuous downscaling of transistors over many technological generations has improved the density and performance of these devices [3], leading to tremendous growth in the microelectronics industry. The realization of complex digital systems on a single chip is enabled by modern very-large-scale integration (VLSI) technology. The high demand for portable electronics in recent years has significantly increased the demand for power-sensitive designs with sophisticated features. Highly advanced and scalable VLSI circuits meet the ever-increasing demand in the electronics industry. Continuous device downscaling is one of the major driving forces of IC technology advancement with improved device performance. Currently, devices are being scaled down to the sub-3-nm-gate regime and beyond.

Aggressive downscaling of CMOS technology has created many challenges for device engineers and new opportunities. The semiconductor process complexity increases as the transistor dimensions decrease. As we approach atomic dimensions, simple scaling eventually stops. Although devices are small, many aspects of their performance deteriorate, e.g., leakage increases [4, 5, 6]; gain decreases; and sensitivity to process variations in manufacturing increases [7]. The profound increase in process variations significantly impacts the circuit operation, leading to a variable performance in identical-sized transistors. It further impacts the propagation delay of the circuit, which behaves as a stochastic random variable, thereby complicating the timing-closure techniques and strongly affecting the chip yield [8]. Increasing process variations in the nanometer regime is one of the major causes of parametric yield loss. Multi-gate field-effect transistors (FETs) [9] are more tolerant to process variations than CMOS transistors. However, their performance parameters are also affected by aggressive scaling [10, 11].

Advanced and affordable design techniques with finer optimization must be adopted in the VLSI design flow to maintain future performance trends in circuits and systems. The turnaround time of a chip depends on the performance of electronic design automation (EDA) tools in overcoming design constraints. The traditional rule-based methodologies in EDA take longer to yield an optimal solution for the set design constraints. In addition, to a certain level, the conventional solutions employed for such tasks are largely manual; thus, they are time-critical and resource intensive, resulting in time-to-market delays. Moreover, once the data are fed back, it is difficult and time-consuming for the designers to understand the underlying functionalities, i.e., the root cause of issues, and apply fixes if required. This difficulty increases under the impact of the process and environmental variations [12, 7].

Artificial intelligence (AI) has provided prominent solutions to many problems in various fields. The principle of AI is based on human intelligence, interpreted in such a way that a machine can easily mimic it and execute tasks of varying complexity. Machine learning (ML) is a subset of AI. The goals of AI/ML are learning, reasoning, predicting, and perceiving. AI/ML can quickly identify the trends and patterns in large volumes of data, enabling users to make relevant decisions. AI/ML algorithms can handle multi-dimensional and multivariate data at high computational speeds. These algorithms continuously gain experience and improve the accuracy and efficiency of their predictions. Further, they facilitate decision-making by optimizing the relevant processes. Considering the numerous advantages of AI/ML algorithms, their applications are endless. Over the last decade, AI/ML strategies have been extensively applied in VLSI design and technology.

∗Corresponding author
deepthi.amuru@research.iiit.ac.in (D. Amuru);
harsha.vardhan.reddy@research.iiit.ac.in (H.V. Vudumula);
pavan.kalyan@students.iiit.ac.in (P.K. Cherupally);
sushanth.reddy@students.iiit.ac.in (S.R. Gurram); amirahmad01@gmail.com (A. Ahmad); andleebiitd@gmail.com (A. Zahra); zia.abbas@iiit.ac.in (Z. Abbas)
ORCID(s): 0000-0003-0793-3244 (D. Amuru)

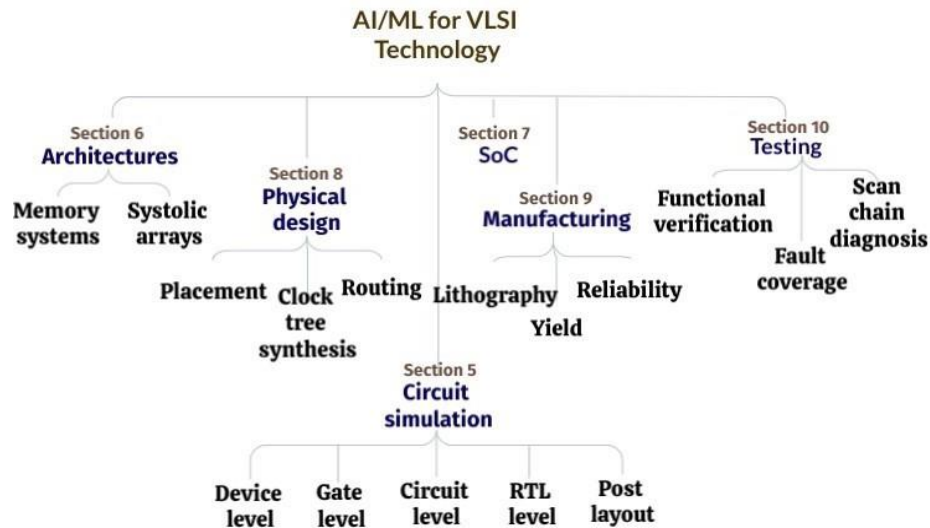

**Figure 1:** Different areas of VLSI Technology reviewed in the paper

VLSI–computer-aided design (CAD) tools are involved in several stages of the chip design flow, from design entry to full-custom layouts. Design and performance evaluation of highly complex digital and analog ICs depends on the CAD tools' capability. Advancement of VLSI–CAD tools is becoming increasingly challenging and complex with the tremendous increase in transistors per chip. Numer- ous opportunities are available in semiconductor and EDA technology for developing/incorporating AI/ML solutions to automate processes at various VLSI design and manufacturing levels for quick convergence [13],[14]. These intelligent learning algorithms are steered and designed to achieve relatively fast turnaround times with efficient, automated solutions for chip fabrication.

This work thoroughly attempts to summarize the literature on AI/ML algorithms for VLSI design and modeling at different abstraction levels. It is the first paper that provides a detailed review encompassing circuit modeling to system-on-chip (SoC) design, along with physical design, testing, and manufacturing. We also briefly present the VLSI design flow and introduction to artificial intelligence for the benefit of the readers.

We organized the paper as follows. Section 2 briefly discusses the existing review articles on AI/ML–VLSI. An overview of artificial intelligence and machine learning and a brief on different steps in the VLSI design and manufacturing are presented in sections 3 and 4, respectively. A detailed survey of AI/ML-CAD-oriented work in circuit simulation at various abstraction levels (device level, gate level, circuit level, register-transfer level (RTL), and post-layout simulation) is presented in Section 5. A review of AI/ML algorithms at the architecture level and SoC level is reported in sections 6 and 7. A survey of the learning strategies proposed in physical design and manufacturing (lithography, reliability analysis, yield prediction, and management) is discussed in sections 8 and 9, respectively. The AI/ML approaches proposed in testing are reported in Section 10. Sources of training data for AI/ML-VLSI are presented in Section 11, followed by challenges and opportunities for AI/ML approaches in the field of VLSI in Section 12.

## 2. Existing Reviews

The impact of AI on VLSI design was first demonstrated in 1985 by Robert. S. Kirk [15]. He briefly explained the scope and necessity for AI techniques in CAD tools at different levels of VLSI design. His paper included a brief on the existing VLSI–AI tools and stressed the importance of incorporating the expanded capabilities of AI in CAD tools. The advantages of incorporating AI in the VLSI design process and its applications are briefed in [16] and [17]. Khan et al. [17] focused on the applications of AI in the IC industry, particularly in expert systems; different knowledge-based systems, such as design automation assistant, design advisor by NCR, and REDESIGN, being used in the VLSI industry. Rapid developments in AI/ML have drawn the attention of researchers who have made numerous pioneering efforts to design, develop, and apply learning strategies to VLSI design and manufacturing. The implementation of neural networks (NNs) for digital and analog VLSI circuits and knowledge-based systems has been reported in [18]. The scope for the joint optimization of physical design with data analytics and ML is reviewed in [19].

Many recent applications and opportunities for ML in physical design are reviewed in [20]. Beerel et al. [21] stated the challenges and opportunities associated with ML-based algorithms in asynchronous CAD/VLSI; they proposed the development of an ML-based recommendation tool, called design advisor, that monitors and records the actions taken by various designers during the usage of standard RTL, logic synthesis, and place route tools. The design advisor chooses the best action for a given scenario by running powerful training engines. Subsequently, the design advisor is deployed and used by circuit designers to obtain design recommendations. Overall, these design advisors focus more on asynchronous CAD/ML tools. Stratigopoulos et al. reviewed IC testing by demonstrating various ML techniques in the field of testing and provided recommendations for future practitioners [22].

Elfadel et al. [23] discussed in detail various ML methods used in the fields of physical design; yield prediction; failure, power, and thermal analysis; and analog design. Khailany et al. [24] highlighted the application of ML in chip designing. They focused on ML-based approaches in micro-architectural design space exploration, power analysis, VLSI physical design, and analog design to optimize the prediction speed and tape-out time. They proposed an AI-driven physical design flow with a deep reinforcement learning (DRL) optimization loop to automatically explore the design space for high-quality physical floorplans, timing constraints, and placements, which can achieve good-quality results, downstream clock-tree synthesis (CTS), and routing steps.

ML in EDA is currently gaining the attention of researchers and research communities. Employing ML in IC design and manufacturing augments the designers by reducing their time and effort in data analysis, optimizing the design flow, and improving time to market [25]. Rapp et al. presented a comprehensive presentation of state of the art on ML for CAD at different abstract levels [26]. Interestingly, the paper also presents a meta-study of ML usage in CAD to capture the overall trend of suitable ML algorithms at various levels of the VLSI cycle. As per the meta-study, the trend for ML-CAD is shifting toward Physical design with NN-implementations compared to other abstraction levels and algorithms. The paper also discusses open challenges while employing ML for CAD, such as the problem of combinatorial optimization, limited availability of training data, and practical limitations. However, the reviews and summaries have been presented only for the last five years, limited to five key conferences and journals. Another survey [27] summarizes ML-CAD works in a well-tabulated manner covering many abstraction levels in digital/analog design flow. However, there needed to be more focus on challenges and future directions. In [28], a comprehensive review of Graphical Neural Networks (GNNs) for EDA is presented, highlighting the areas of logic synthesis, physical design, and verification. As graphs are an intuitive way of representing circuits, netlists, and layout, GNN can easily fit into EDA to solve combinatorial optimization problems at various levels and improve the QoR (Quality of Results) [29]. A review of ML achievements in placement and routing with benchmark results on benchmark ISPD 2015 datasets is presented in [30].

Recently, a brief review of recent machine learning and deep learning techniques incorporated in analog and digital VLSI, including physical design, is discussed in [31]. VLSI Computer-Aided Design at different abstraction levels from a machine-learning perspective is presented in [32]. In [33], applications, opportunities, and challenges of reinforcement learning to EDA, mainly macro chip placement, analog transistor sizing, and logic synthesis, are discussed with practical implementations.

The reviews mentioned above break down to provide a detailed discussion of the AI/ML approaches proposed in the literature, mainly covering all the abstraction levels of the digital VLSI design flow. This review summarizes the literature on AI/ML algorithms for VLSI design and modeling at different abstraction levels. We also discuss the challenges, opportunities, and scope for incorporating automated learning strategies at various levels in the semiconductor design flow. The design abstraction levels covered in this review under different sections are shown through a dendrogram in fig.1. A concise VLSI design flow with the traditional commercial CAD tools used in the industry and the surrogate AI/ML techniques proposed by researchers is given in fig.2. Figure 6 provides a summary of the AI/ML techniques proposed in the literature for VLSI circuit simulation for estimating circuit performance parameters, such as the transistor characteristics, statistical static timing analysis (SSTA), leakage power, power consumption, and post-layout behavior.

In the following sections, we present a brief background of AI/ML and a brief description of the different stages of the VLSI design flow.

## 3. Brief on VLSI Design Flow

A traditional digital IC design flow has many hierarchical levels, as shown in fig.2; the flowchart covers a generalized design flow, including the front-end and back-end of full-custom/semi-custom IC designs. The design specifications abstractly describe the functionality, interface, and overall architecture of the digital circuit to be designed. They include block diagrams providing the functional description, timing specifications, propagation delays, package type required, and design constraints. They also act as an agreement between the design engineer and vendor. The architectural design level comprises the system's basic architecture. It includes decisions such as reduced instruction set computing/complex instruction set computing (RISC/CISC) processors and the number of arithmetic logic units (ALUs) and floating-point units. The outcome of this level is a micro-architectural specification that contains the functional descriptions of subsystem units. Architects can estimate the design performance and power based on such descriptions.

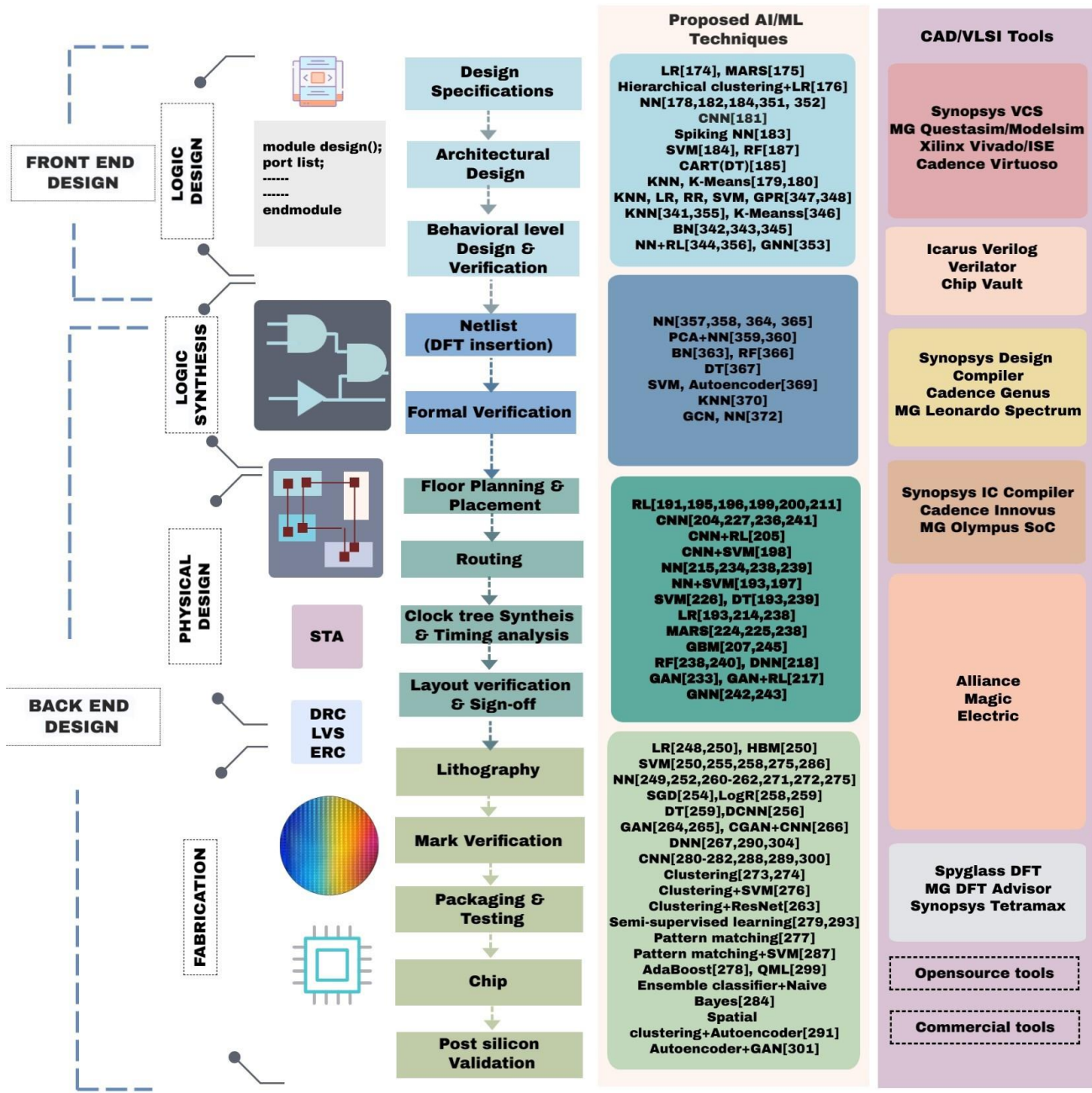

**Figure 2:** Modern Chip Design Flow

The behavioral design level is the next; it provides the functional description of the design, often written using Verilog HDL/VHDL. The behavioral level comprises a high-level description of the functionality, hiding the underlying implementation details. The timing information is checked and validated in the next level, i.e., the RTL description (register transfer level). A high-level synthesis (HLS) tool can automatically convert C/C++-based system specifications to HDL. Alternatively, the logic synthesis tool produces the netlist, i.e., a gate-level description for the high-level behavioral description. The logic synthesis tool ensures that the gate-level netlist meets the timing, area, and power specifications. Logic verification is performed through testbench/simulation. Formal verification and scan insertion through design for testability (DFT) are performed at this stage to examine the RTL mapping [34]. Next, system partitioning, which divides large and complex systems into small modules, is performed, followed by floor planning, placement, and routing. The primary function of the floor planner is to estimate the required chip area for standard cell/module design implementation and is responsible for improving the design performance. The place and route tool place the sub-modules, gates, and flip-flops, followed by CTS (clock tree synthesis) and reset routing. Subsequently, the routing of each block is performed. After placement and routing, layout verification is performed to determine if the designed layout conforms to the electrical/physical design rules and source schematic. These processes are implemented using tools such as design rule check (DRC) and electrical rule check (ERC). After the post-layout simulation, where parasitic resistance and capacitance extraction and verification are performed, the chip moves to the sign-off stage [35]. GDS-II is the resultant file sent to the semiconductor foundries for IC fabrication.

IC fabrication involves many advanced and complex physical and chemical processes that must be performed with utmost precision. It comprises numerous stages, from wafer preparation to reliability testing. A detailed description of each stage is presented in [36]. In brief, silicon crystals are grown and sliced to produce wafers. The wafers must be polished to near perfection to achieve extremely small dimensions of VLSI devices. The fabrication process comprises several steps, including the deposition and diffusion of various materials on the wafer. The layout data from the GDS-II file is converted into photolithographic masks, one for each layer. The masks define the spaces on the wafer where certain materials need to be deposited, diffused, or even removed. During each step, one mask is used. Several dozen masks may be used to complete the fabrication process. Lithography is the step that involves mask preparation and verification as well as the definition of different materials in specific areas of the IC. It is a crucial step during fabrication and is repeated numerous times at different stages. It is the step most affected by the downscaling of technology nodes and the increase in process variations. After the chip is fabricated, the wafer is diced, and individual chips are separated. Subsequently, each chip is packaged and tested to validate the design specifications and functional behavior. Post-silicon validation is the last step in IC manufacturing and is used to detect and fix bugs in ICs and systems after production [37].

## 4. Brief on AI/ML algorithms

In modern times, statistical learning plays a crucial role in nearly every emerging field of science and technology. The vast amount of data generated and communicated within each field can be mined for learning patterns and dependencies among the parameters for future analyses and predictions. The statistical learning approach can be applied to solve many real-world problems. AI is a technology that enables a machine to simulate human behavior. ML and deep learning are the two main subsets of AI. ML allows a machine to automatically learn from past data without explicit programming. Deep learning is the prime subset of ML (Fig. 3(a)). ML includes learning and self-correction when new data is introduced. ML can handle structured and semi-structured data, whereas AI can handle structured, semi-structured, and unstructured data. ML can be divided into three main types: supervised, unsupervised, and reinforcement learning. Supervised learning is performed when the output label is present for every element in the given data. Unsupervised learning is performed when only input variables are present in the data. The learning that involves data with a few labeled samples and the rest is unlabeled is referred to as semi-supervised learning [38].

### 4.1. Supervised Learning

Supervised learning is further divided into two classes: classification and regression. Classification is a form of data analysis that extracts models describing important data classes. Such models, called classifiers, predict discrete categorical class labels [39]. In contrast, regression is used to predict missing or unavailable numerical data rather than discrete class labels. Regression analysis is a statistical methodology generally used for the numeric prediction of continuous-valued functions [40]. The term prediction refers to both numeric and class-label predictions. The classification/regression process can be viewed as a learning function to predict a mapping of $Y = f(X)$ where $Y$ is a set of output variables for $X$ input variables. The mapping function is estimated for predicting the associated class label y of a given new tuple $X$ (Fig. 3(b)). The most considerable drawback of supervised learning is that it requires a massive amount of unbiased labeled training data, which is hard to produce in specific applications such as VLSI. Most popular regression and classification algorithms include linear, polynomial, and ridge regressions; decision trees (DT); random forest (RF); support vector machines (SVMs); and ensembled learning [41, 42].

### 4.2. Unsupervised Learning

In contrast to supervised learning, unsupervised learning does not require a label for each training tuple. Hence, it requires less effort to generate the data than supervised learning. However, point estimates/desired output for a required input vector is harder to achieve with unsupervised learning. It is employed to identify unknown patterns in the data. Clustering and dimensionality reduction through principal component analysis and other methods are powerful applications associated with unsupervised learning. Clustering involves grouping or segmenting objects into subsets or "clusters" such that the objects in each cluster are more closely related to one another than to the objects of different clusters. For a more detailed discussion, refer to [43]. Common clustering algorithms include K-nearest neighbors (KNN), K-means clustering, hierarchical Clustering, and agglomerative clustering [44].

### 4.3. Semi-supervised Learning

Semi-supervised learning acts as a bridge between supervised and unsupervised methodologies. It is useful when training data has limited labeled samples and a large set of unlabeled samples. It works great to automate data labeling.

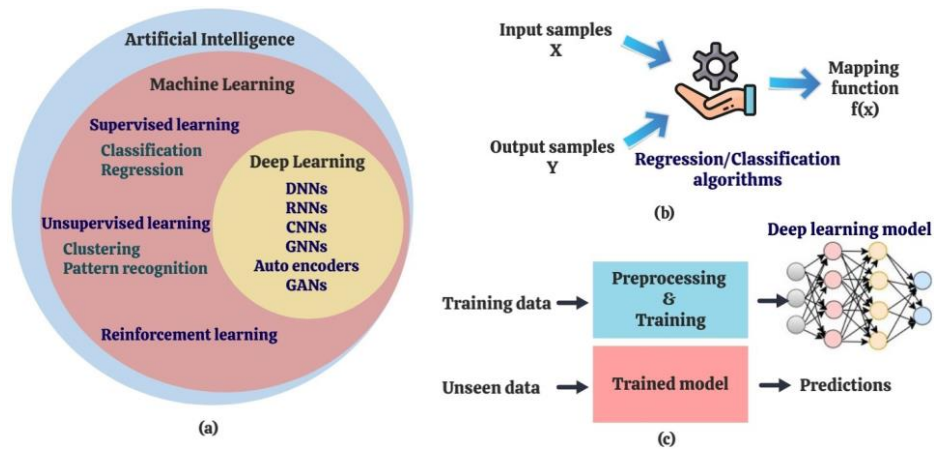

**Figure 3:** (a) Overview of Artificial Intelligence techniques (b) Learning function of classification/regression algorithms (c) Deep learning training and prediction

It works better than supervised/unsupervised learning alone in some applications. The training starts with limited labeled data and then applies algorithms to model the unlabeled dataset with pseudo labels in the next step. Then, the labeled data is linked with pseudo labels and later with unlabeled data to improve the accuracy [45, 46]. However, many efforts are needed to converge both parts of the semi-supervised methodology in certain complex applications.

### 4.4. Reinforcement Learning

Reinforcement learning is an area of machine learning that maps the situations to actions to maximize a numerical reward signal; it is focused on goal-directed learning based on interactions [47]. It does not rely on examples of correct behavior as in the case of supervised learning or does not try to find a hidden pattern as in unsupervised learning. Reinforcement learning is trying to learn from experience and find an optimum solution that maximizes a reward signal.

### 4.5. Deep Learning

Deep learning is a subset of ML and is particularly suitable for big-data processing. Deep learning enables the computer to build complex concepts from simple concepts [48]. A feed-forward network or multi-layer perceptron (MLP) is an essential example of a deep learning model or artificial neural network (ANN) (Fig. 3(c)). An MLP is a mathematical function mapping a set of input and output values. The function is formed by composing many simple functions. A shallow neural network (SNN) is an NN with one or two hidden layers. A network with tens to hundreds of such layers is called a deep neural network (DNN). DNNs extract features layer by layer and combines low-level features to form high-level features; thus, they can be used to find distributed expressions of data [49]. Compared with shallow neural networks (SNNs), DNNs have better feature expression and the ability to model complex mapping. Frequently used DNNs include deep belief networks, stacked autoencoder (SAE), and deep convolution NNs (DCNNs) [48]. Recently, DNNs have revolutionized the field of computer vision. DCNNs are suitable for computer vision tasks [50]. Other popular deep learning techniques include recurrent NNs (RNNs) [51]; generative adversarial networks (GANs) [52, 53]; and DRL (deep reinforcement learning) [54]. Refer to the research works mentioned in fig. 2 for implementation details of these algorithms.

Rapid development in several fields of AI/ML is increasing the scope for solution creation to address many divergent problems associated with IC design and manufacturing. In the following sections, we discuss the applications of AI/ML at different abstraction levels of VLSI design and analysis, starting with circuit simulation.

## 5. AI at the Circuit Simulation

Simulation plays a vital role in IC device modeling. Performance evaluation of designed circuits through simulations is becoming quite challenging in the nanometer regime due to increasing process and environmental variations [55, 56, 57]. The ability to discover functional and electrical performance variations early in the design cycle can improve the IC yield, which depends on the simulation tools' capability. By assimilating the automated learning capabilities offered by AI/ML algorithms in E-CAD tools, the turnaround time and performance of the chip can be revamped with reduced design effort. Researchers have proposed surrogate methodologies targeting the characterization of the leakage power, total power, dynamic power, propagation delay, and IR-drop estimation ranging from stack-level transistor models to the subsystem level [58]. Different AI/ML algorithms have been explored for circuit modeling at different abstraction levels, including linear regression (LR), polynomial regression (PR), response surface modeling (RSM), SVM, ensembled techniques, Bayes theorem, ANNs, and pattern recognition models [59]. The following subsections describe the learning strategies proposed in the literature for VLSI device/circuit characterization at different abstraction levels.

## 5.1. DEVICE LEVEL

Parametric yield estimation of the circuit and device modeling at the transistor level is the primary focus area at this level. Parametric yield estimation of statistical-aware VLSI circuits is not new; this process has been evolving along with ML algorithms since the 1980s. Statistical parametric yield estimation was proposed [60] for determining the overall parametric yield of MOS circuits. Alvarez et al. and Young et al. proposed a statistical design analysis through a response surface methodology (RSM) for computer-aided VLSI device design [61, 62]. The proposed models have been successfully applied to optimize the BiCMOS transistor design. RSM has inspired industrial experimentation since its development in the 1950s. Refer to [63],[64] for a comprehensive review of RSM. Khan et al. [65] proposed the multivariate polynomial regression (MPR) method for approximating the early voltage and MOSFET characteristics in saturation; they considered a curve-fitting approach using the least-squares method in MPR for simplifying the complexity in BSIM3, and BSIM4 equations [66] to calculate the MOSFET characteristics realistically.

Considering the drastic decrease in the dimensions of technology nodes, conducting a thorough analysis of the characteristics at the device level is of utmost necessity. The randomness in the behavior of transistors due to the inter-die and intra-die variations in the process causes exponential changes in the device currents, particularly in the sub-threshold [56]. Statistical sampling techniques are more effective than conventional corner-based methods for estimating the effect of the process parameters on the device [67]. The datasets generated from the statistical sampling techniques are best suited for learning strategies. The development of AI/ML algorithms for analyzing device parameters at different technology nodes facilitates the optimization of the device parameters and estimating the parametric yield at very high computational speeds. Owing to this fact, an ML-based Tikhonov regularization (TR) approach is implemented to analyze the impact of the process on $V_{TH}$ in GaN-based high electron mobility transistors (HEMTs) [68]. In [69], neural network-based variability analysis of ferroelectric field-effect transistor (FeFET) with raw data in the form of polarization maps from the metrology as inputs is proposed. High/low threshold voltage, on-state current, and sub-threshold slope are sampled as outputs from the model. The experiments show that ML predictions are $10^6$ times faster and > 98% accurate compared to TCAD simulations. A hybrid analytical and deep-learning-assisted MOSFET I-V (current-voltage) modeling is proposed in [70]. For modeling the I-V characteristics of a 12nm gate length GAAFET (Gate-all-around transistor) technology, a 3-layer NN with 18 neurons was employed.

Performance evaluation of FinFET devices and circuits designed at 7 nm and above is becoming challenging. Accurate estimation of the reliability of these devices prior to manufacturing is another concern [71]. Identifying the trend in ML applications for device modeling from RSM to ANNs over the years and noticing the future requirements in advanced technologies, we propose inductive transfer learning [72, 73] as a promising technique for investigating the device behavior in forthcoming technology nodes from the knowledge of existing technology nodes.

Given a source domain, $D_S$, a corresponding source task, $T_S$, a target domain, $D_T$, and a target task, $T_T$, the objective of transfer learning is to enable the learning of the target conditional probability distribution, $P(Y_T|X_T)$ in $D_T$ with the information gained from $D_S$ and $T_S$ where $D_S \neq D_T$ or $T_S \neq T_T$. In most cases, a limited number of labeled target examples are assumed to be available, exponentially smaller than the number of labeled source examples. Fig. 4 shows the proposed methodology for developing a learning system using transfer learning to analyze the behavior of devices in upcoming technology nodes.

## 5.2. GATE LEVEL

Researchers have explored the application and development of AI/ML techniques for gate-level circuit design and evaluation. Figure 5 shows generalized modeling of statistical aware circuit simulation at the gate level. Down the line, RSM modeling was popular for estimating process variation effects on the circuit design. Mutlu et al. presented a detailed analysis of the development of RSMs to estimate the process variation effects on the circuit design [74]. Basu et al. [75] developed a library of statistical intra-gate variation tolerant cells by building RSM-based gate-delay models with reduced dimensions; the developed, optimized standard cells can be used for chip-level optimization to realize the timing of critical paths. In [76], [77] RSM learning models were developed via a combination of statistical design of experiment (DoE) and an automatic selection algorithm for the SSTA of the gate-level library-cell characterization of VLSI circuits. Their models considered the threshold voltage ($V_{th}$) and current gain ($\beta$) as model parameters for a compact transistor model characterization of power, delay, and output transitions. In [76], the RSM and linear sensitivity approaches were proposed to increase the analysis speed by one and two orders of magnitude, respectively, when compared to that of Monte Carlo (MC) simulations, albeit at the cost of a decrease in accuracy of up to 2% and 7% respectively. In [77], on average, s-DoE has an error of 0.22% at the tails of $3\sigma$ distribution compared to the 10x error given by sensitivity analysis by cadence encounter library characterizer (ELC).

Miranda et al. [78] also proposed a variation-aware statistical design of experiments approach (s-DoE) for predicting the parametric yield of static random access memory (SRAM) circuits under process variability. Their approach achieved an accuracy of approximately two orders of magnitude better than that for the sensitivity analysis in the tail response under $3\sigma$ process variations and a CPU time 10–100 times less than that in MC simulations. The case studies in the article demonstrate the advantage of s-DoE in choosing the region of interest in the distribution to improve accuracy while reducing the number of simulations. Under similar lines, Chaudhuri et al. [79] developed accurate RSM-based analytical leakage models for 22nm shorted-gate and independent-gate FinFETs using a central composite rotatable design to estimate the leakage current in FinFET standard cells by considering the process variations. Their results agreed well with the quasi-MC simulations performed in TCAD using 2D cross-sections.

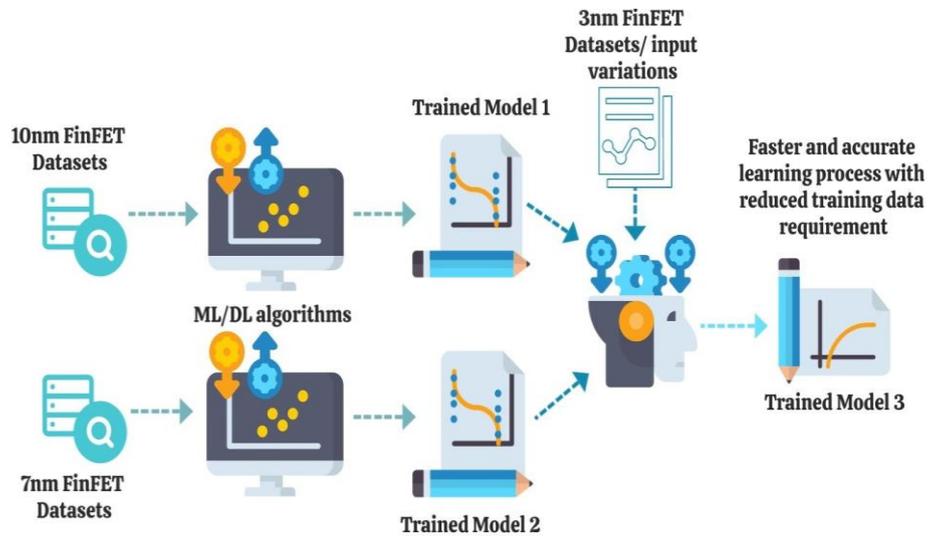

**Figure 4:** Block diagram of the proposed inductive transfer learning for device modeling at upcoming lower technology nodes

Exploration of possible patterns in simulated data and reuse of the data across various stages of circuit design was of great interest. In this fashion, Cao et al. [80] proposed a robust table-lookup method for estimating the gate-level circuit leakage power and switching energy of all possible states using the Bayesian interface (BI) and neural networks (NNs). Their model uses pattern recognition by classifying the possible states based on the average power consumption values using NNs. The idea is centered on using the statistical information on a circuit's available SPICE power data points to characterize the correlation between the state-transition patterns and power consumption values of the circuit. Such correlated pattern information is further utilized to predict the power consumption of any seen and unforeseen state transition in the entire state-transition space of the circuit. The estimation errors obtained using NNs always exhibit normal distributions, with much smaller variations than benchmark curves. Moreover, the estimation error decreases with the number of clusters and complexity of the NNs when appropriate features are extracted. Additionally, the time required to train and validate the NNs is negligible compared to the computing time required to generate statistical distributions using the SPICE environment.

Applying BI, Yu et al. [81] proposed a novel nonlinear analytical timing model for statistical characterization of the delay and slew of standard library cells in bulk silicon, SOI technologies, and non-FinFET and FinFET technologies, using a limited combination of output capacitance, input slew rate, and supply voltage. Utilizing the Bayesian inference framework, they extract the new timing model parameters using an ultra-small set of additional timing measurements from the target technology, achieving a 15× runtime speedup in simulation runs without compromising accuracy, which is better than the traditional lookup table approach. They employed ML to develop priors of timing model coefficients using old libraries and sparse sampling to provide the additional data points required for building the new library in the target technology.

Over time, polynomial regression was another important analytical modeling technique. A statistical leakage estimation through PR was proposed by [82]. Experimental results on the MCNC benchmark [83] show that the leakage estimation is five times more efficient than Wilkinson's approach [84] with no accuracy loss in mean estimation and about 1% in standard deviation. On these lines, Moshrefi et al. [85] proposed an accurate, low-cost Burr distribution as a function for delay estimation at varying threshold voltages ±10% from mean. The samples are generated at the 90, 45, and 22nm technology nodes. Statistical data from MATLAB were applied to HSPICE for simulations to obtain delay variations. The relation between the threshold voltage and delay variations was determined as a fourth-order polynomial equation. In addition to the mean and variance of the estimated distributions, the maximum likelihood was considered the third parameter, forming a three-parameter probability density function. The proposed Burr distribution benefits with one more degree of freedom to the normal distribution [86], and with lower error distribution.

The AI/ML predictive algorithms are intermittently applied for the process–voltage–temperature (PVT) variation-aware library-cell characterization of digital circuit design and simulation. The accurate performance modeling of digital circuits is becoming difficult with the acute downscaling of transistor dimensions in the deep sub-micrometer regime [87], [88]. To address the concern regarding the performance modeling of digital circuits in the sub-micrometer regime, Stillmaker et al. [89] developed polynomial equations for curve-fitting the measurements of the CMOS circuit delay, power, and energy dissipation based on HSPICE simulated data using predictive technology models (PTMs) [90] at technology nodes ranging from 180 nm to 7 nm. Second-order and third-order polynomial models were developed with iterative power, delay, and energy measurement experiments, attaining a coefficient of determination (R2score [91]) of 0.95. The scaling models proposed in [92] and [89] are more accurate for comparing devices at different technology nodes and supply voltages than the classical scaling methods.

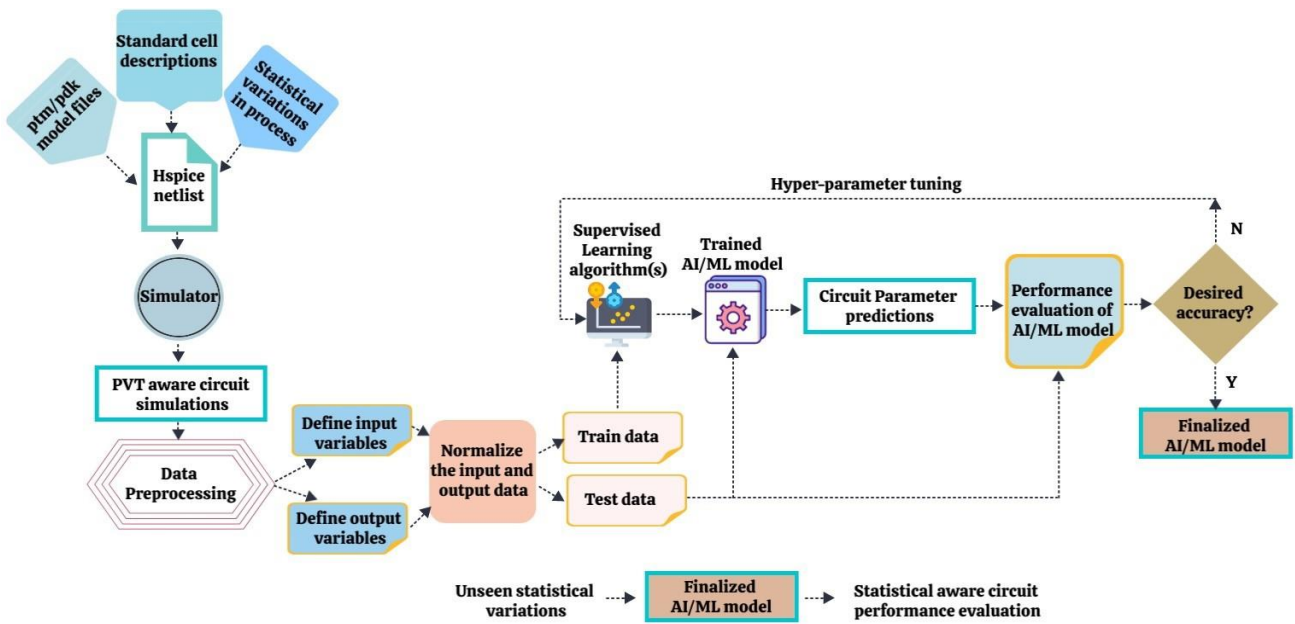

**Figure 5:** Generalized statistical aware modeling for VLSI circuit simulation

Development of MPR and ANN models for measuring the PVT-aware (process voltage temperature) leakage in CMOS and FinFET digital logic cells was reported in [91], and [93] respectively. [91] also models total power with the same MPR model. The developed models demonstrated high accuracy with < 1% error w.r.t. the HSPICE simulations. Amuru et al. [94] reported a PVT-aware estimation of leakage power and propagation delay with a Gradient boosting algorithm, which yields a < 1% error in estimations with $10^4$ times improvement in computational speed compared to HSPICE simulations. These characterized library-cell estimations can be used for estimating the overall leakage power and propagation delay of complex circuits, avoiding the relatively long simulation runs of traditional compilers. Bhavesh et al. [95] propose an estimation of power consumption of the MOSFET-based digital circuits using regression algorithms. PMOS-based Resistive Load Inverter (RLI), NMOS-based RLI, and CMOS-based NAND gate layout are employed at 90nm MOS technology to create the dataset. The feature vectors extracted for modeling are capacitance, resistance, number of MOSFET, their respective width and length, and the average power consumption of the respective layout. As per the experimental results, Extra tree and polynomial regressors demonstrate better performance over Linear, RF, and DT regressors. GPU-based circuit analysis is required at the present state of complex-circuit analysis. Recently, XT-PRAGGMA, a tool to eliminate false aggressors and accurately predict crosstalk-induced delta delays using GPU-accelerated dynamic gate-level simulations and machine learning, is proposed in [96]. It shows a speedup of 1800x compared to SPICE-based simulations.

An accurate yield estimation in the early stage of the design cycle can positively impact the cost and quality of IC manufacturing [97],[98]. Comprehensive analysis of VLSI circuits' delay and power characteristics being designed at the sub-nanometer scale under expanding PVT variations is extremely important for parametric yield estimation. As reported earlier, accurate predictions are made by well-trained AI/ML algorithms, such as PR, ANNs, GB, and BI, with power and delay estimations that are very close to those of the most-reliable HSPICE models. Incorporating such efficient ML models in EDA tools for library-cell characterization at the transistor level and gate level facilitates the performance evaluation of complex VLSI circuits at very high computational speeds, facilitating the analysis of the yield. These advanced computing EDA tools drastically improve the turnaround time of the IC.

### 5.3. CIRCUIT LEVEL

Statistical characterization of VLSI circuits under process variations is essential for avoiding silicon re-spins. Similar to gate-level, explorations for the design of ML-based surrogate models at the circuit level were reported in the literature. Hou et al. [99] reported the power estimation of VLSI circuits using NNs. Trained NNs can estimate the power using the input/output (I/O) and cell number without requiring circuit information such as net structures. This approach requires the power estimation results of benchmark circuits to train the target NN. Limited experimental results have shown that this method can give acceptable results with a specific net structure at a considerably high speed. Stockman et al. [100] discussed a novel approach for predicting power consumption based on memory activity counters, exploiting the statistical relationship between power consumption and potential variables. The proposed ML models for the prediction include support vector regression (SVR), genetic algorithms, and NNs. They showed that a NN with two hidden layers and five nodes per layer is the best predictor among the chosen ML models, with a mean square error of 0.047. In addition, they explained that the ML approaches are significantly less costly and less complex than a hardware solution, with reduced run time. Janakiraman et al. [101] proposed an efficient ANN model for characterizing the voltage- and temperature-aware statistical analysis of leakage power. Trained transistor-level stack models used for circuit leakage estimation. The designed model showed 100x improvement in runtime with < 1% and < 2% error in the mean and standard deviation of Monte-Carlo statistical leakage estimations. The complexity of the comprehensive model is reported as $O(N)$ on par with existing linear and quadratic models [102, 103, 84, 104].

Garg et al. presented SVM-based macro models for characterizing transistor stacks of CMOS gates with an average increase in the runtime of 17× compared to those of the HSPICE computations for estimating the leakage power [105]. Kahng et al. [106] proposed a hybrid surrogate model that combines the predictions of ANN and SVM models to estimate the incremental delay due to the signal integrity aware path delay in a 28-nm FDSOI technology, demonstrating a worst-case error of $< 10ps$. An accurate power estimation of CMOS VLSI circuit using Random Forest (RF) that performs better than NNs is proposed in [107]. Results show a good agreement with ISCAS'89 Benchmark circuits. A fast and efficient ResNet-based digital circuit optimization framework for leakage and delay is proposed in [108]. Results on 22nm Metal Gate High-K digital cells show 36.7% and 18.8% reduction in delay and leakage using a genetic algorithm.

### 5.4. RTL level

The effect of process variability on guard bands and its mitigation are detailed in [109]. Jiao et al. [110] proposed a supervised-learning model for the bit-level static timing error prediction modeled at the RTL level, aiming for a guard-band reduction in error-resilient applications. They considered floating-point pipelined circuits in their analysis. The circuit's behavior was characterized by timing errors using the Synopsys design and Synopsys IC compilers as frontend and backend design tools, respectively. Synopsys prime-time was used for voltage and temperature scaling, followed by a post-layout simulation with the SDF back-annotation in Mentor Graphics ModelSim to extract the bit-level timing error information. The logistic regression model shows an average accuracy of 95% at various voltage/temperature corners and unseen workload, with an average guard-band reduction of 10%. ML-based power estimation techniques at the RTL level that outperform commercial RTL tools [111, 112, 113, 114] were proposed in [115]. Their experiments recommend CNN over ridge regression, gradient tree boosting, and multi-layer perceptron for accurate power estimations. The average power estimation from the RTL simulations using a GNN [116], GRANNITE, was presented in [117]. GRANNITE achieved $> 18.7X$ speedup when compared to traditional per-cycle gate-level simulations.

The AI/ML strategies can be extended to the circuit and RTL level to build macrocell models for parametric yield estimation and optimization. The models built using ANNs, CNNs, and deep learning techniques are helpful for complex cell design optimization and power-delay product prediction as they are less dependent on the complete circuit description. Another critical bottleneck is the generation of big data for ML algorithms. ML algorithms require a large amount of simulated data to accurately develop I/O relationships, which is possible at some levels of digital circuits and their applications. The concept of GANs can help address this concern. Generative models aim to estimate the training data's probability distribution and generate samples belonging to the same data distribution manifold [118]. GAN-based semi-supervised method architectures for the regression task proposed recently [119] strengthen the possibilities of applying GAN to the regression tasks of digital circuits. Different measures and techniques need to be explored to keep the quantization error introduced by these networks in check.

### 5.5. POST LAYOUT SIMULATION

ML models also facilitate the efficient use of resources in repeated dynamic IR-drop simulations. The model proposed in [120] reduces the training time by building small-region models for cell instances for IR-drop violations instead of building a global model for the entire chip. Further, ML models work on the regional clusters to extract the required features and predict the violations. Experiments on validated industry designs show that the XGBoost model outperforms CNNs for IR-drop prediction, requiring less than 2 min for each ECO iteration. Zhiyao Xie et al. [121] developed a fast design independent dynamic IR-drop estimation technique named PowerNet based on CNNs. Design-dependent, ML-based IR-drop estimation techniques are proposed in [120, 122, 123, 124, 125].

Han et al. [126] proposed an ML-based tool called Golden Timer eXtension (GTX) for sign-off timing analysis. Using the proposed tool, they attempted to predict the timing slack between different timing tools and the correlation between the sign-off tool and implementation tool across multiple technology nodes. The poor yield due to the inaccurate timing estimation by the STA sign-off, particularly at nodes below 16 nm at low voltages, can be improved using surrogate tools, supporting advanced processes for accurate timing calibration. ML techniques in chip design and manufacturing, notably addressing the effect of process variations on chip manufacturing at the sub-22-nm regime, are discussed in [127]. The authors discuss pattern-matching techniques integrated with ML techniques for pre-silicon HD, post-silicon variation extraction, bug localization, and learning techniques for post-silicon time tuning. [128] reviews some of the on-chip power grid design solutions using AI/ML approaches. It thoroughly discusses Power grid analysis using probabilistic, heuristic, and machine-learning approaches. It further recommends that it is necessary to obtain the electromigration-aware aging prediction of the power grid networks during the design phase itself.

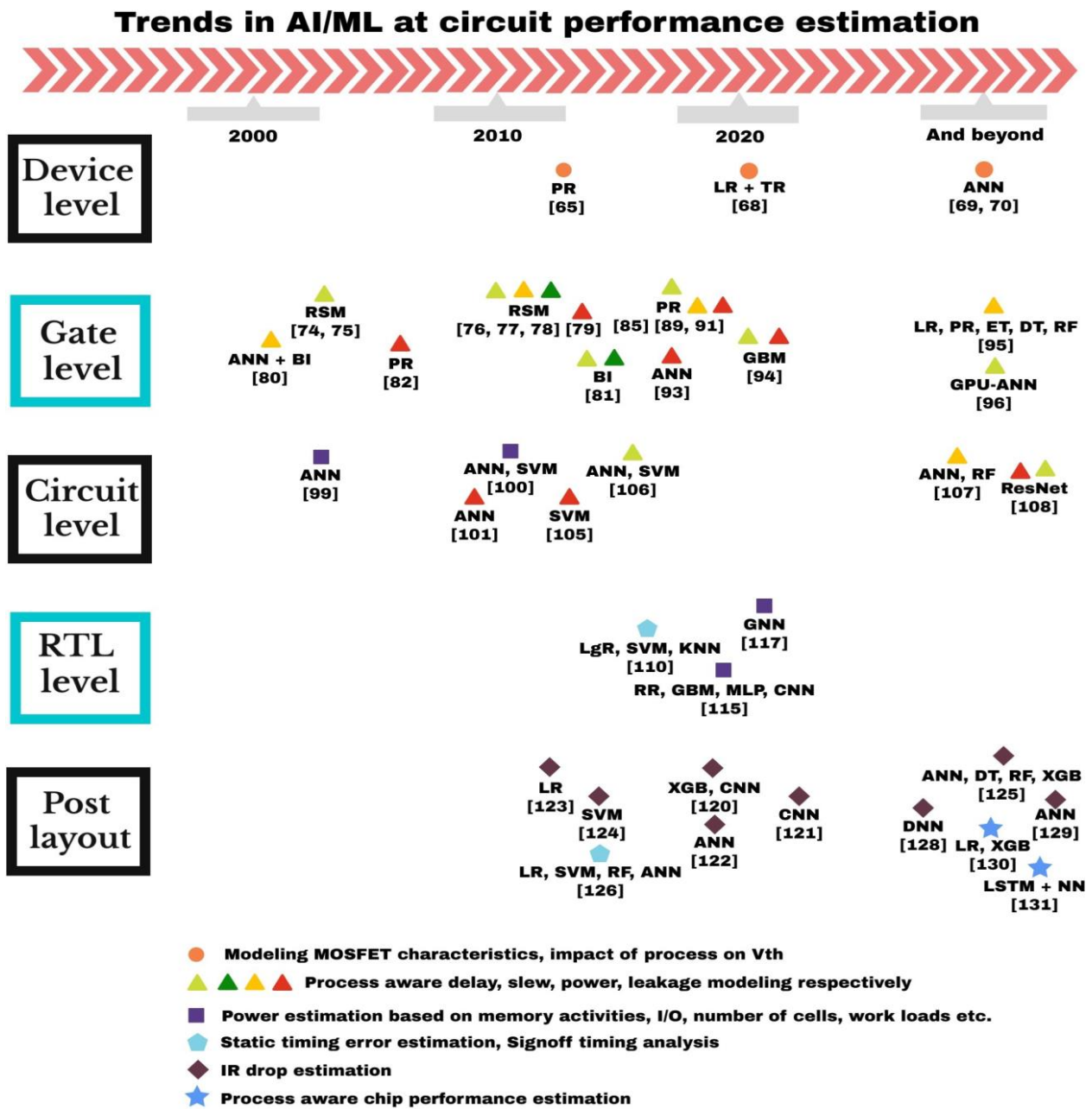

**Figure 6:** Summary of proposed AI/ML algorithms in literature for circuit simulation parameter estimation/performance evaluation

Power delivery networks (PDNs) supply low-noise power to the active components of the ICs. As the supply volt- age scaled down, the variations in power supply voltage increased, affecting the system's performance, especially at higher frequencies. The effects of this power supply noise can be minimized with a proper design of impedance-controlled PDN. The probability of system failure increases with the PDN ratio (The ratio of the actual impedance of the PDN to the target impedance). It can be minimized by efficiently selecting and placing decoupling capacitors on the board and/or the package. A fast ML-based surrogate-assisted meta-heuristic optimization framework for decoupling capacitor optimization is proposed in [129].

Further, a low-cost machine learning-based chip performance prediction framework using on-chip resources is proposed [130]. It predicts the maximum operating frequency of chips for speed binning with an accuracy of over 90% w.r.t Automatic Test equipment (ATE). Experimental results on $12nm$ industrial chips show that linear regression is more suitable with less training time and model size than XGBoost. It also proposes a sensor selection method to minimize the area overhead on on-chip sensors. Sadiqbatcha et al. [131] propose RealMaps, a framework for real-time estimation of full-chip heatmaps using an LSTM-NN (Long short-term memory) model with existing embedded temperature sensors and system-level utilization information. The experiments to identify the dominant spatial features through 2D spatial DCT (Discrete Cosine Transform) shows that only 36 DCT coefficients are required to maintain sufficient accuracy. Fig 6 presents a summary of the AI/ML algorithms proposed in the literature to address VLSI circuit simulation.

As reported earlier, AI and ML can be incorporated into EDA tools and methodologies at various stages of circuit simulation to address different statistical/parameter estimations, including the leakage power, total power, propagation delay, and effects induced due to aging, yield, and power consumption. Assimilation of these automated learning strategies into VLSI circuit design and simulation will revolutionize the field of CAD–VLSI considering the numerous related advantages.

## 6. AI in Architectures

Design of VLSI architectures became dynamic with the evolution of AI/ML techniques [132, 133]. Advances in NN algorithms and innovations in high bandwidth and high-performance semiconductor designs have paved a new way to address the challenges in hardware implementations of advanced real-time applications. Over the last few decades, different architectures have inspired the advancement of VLSI technology. Most design developments/improvements are motivated by the need for edge applications with high processing speeds, improved reliability, low implementation cost, and time-to-market windows. The architectural designs proposed in the literature are majorly for the application domains of image processing and signal processing, speech processing, IoT, and automobile.

This survey presents a broad review of VLSI architectural modifications at the memory and systolic array architectures in this section and at the SoC level in the next section to provide the authors with an overview and scope of research in VLSI architectures for ML.

### 6.1. Memory Systems

Memory systems are one of the computational systems' essential and dominant components. Different scalable memory architectures have been designed for the real-time processing of ML algorithms in various IoT (Internet of Things) and embedded system applications. Various AI applications involve large datasets and demand a faster interface between the computing unit and memory. Different memory architectures were proposed in the past, addressing data movement and processing issues. Kang et al. [134] proposed deep embedding of computation in an SRAM parallel processing architecture for pattern recognition in $256 \times 256$ images; their model enables multi-row read access and analog signal processing without degrading the system performance. Their method employs two models: multi-row READ and the analog sum of absolute difference (SAD) computation. This architecture differs from conventional architecture as a data path between the processor and memory is not required. The SAD is computed at different locations of the array in parallel with multiple windows of the template pattern. For high-performance computations, Zhang et al. [132] proposed a 6T-SRAM array that stores an ML classifier model, which is an ultra-low energy detector for image classification. The prototype is a $128 \times 128$ SRAM array that operates at 300 MHz, with an accuracy equivalent to that of a discrete SRAM/digital-MAC system.

Gonugondla et al. [135] presented a robust, deep-in-memory ML classifier with a stochastic gradient descent based on an on-chip trainer using a standard 16 kB 6T-SRAM bit-cell array. In-memory computing is a technology that uses memory devices assembled in an array to execute MAC operations [136]. Kang et al. [133] worked on deep in-memory architecture (DIMA) as a substitute for the regular von Neuman architecture for realizing energy and latency-efficient ML SoCs. This architecture was employed mainly for targeted applications, such as IoT and autonomous driving, which require computing heavy ML algorithms. DIMA eliminates the need for separate computation and memory by implanting the conventional memory periphery with the computation hardware. The design employs 6T SRAM with a changeless bit-cell structure to maintain the storage density. In [137], MAC circuit architecture in a 2T–1C configuration (two MoS2 FETs and one metal-insulator-metal capacitor) is the core module for the convolution operation in an artificial neural network. The memory portion of this circuit is similar to Dynamic Random Access Memory (DRAM) but with a longer retention time owing to the ultralow leakage current of the MoS2 transistors.

Wang et al. [138] discussed parallel digital VLSI architecture for combined SVM training and classification. In this parallel architecture, a multi-layer system bus and multiple distributed memories fully utilize parallelism. Before this, many SVMs were developed and discussed in [139, 140, 141], primarily focusing on the 90-nm technology node. Distinctively, Wang et al. in [138] developed the SVM on a 45-nm node on a commercial GPU with an enhanced speedup of 29× compared with traditional SVMs on digital hardware.

A part of the computation tasks can be performed inside the memory to solve the data movement issue, thus avoiding the memory access bottleneck and accelerating the application performance significantly. Such architectures are processing in-memory (PIM) architectures. [142] proposes NNPIM, a novel PIM architecture, to accelerate NN's interface inside the memory. The memory architecture combines crossbar memory architecture for faster operations, optimization techniques to improve the NN performance and reduce energy consumption, and weight sharing mechanism to reduce the computational requirement of NNs. [143, 144] are some of the significant state-of-the-art DRAM PIM architectures.

Another evolved computing technique is near-memory processing (NMP). Near-memory processing incorporates the memory and logic chips in 3D storage packages to provide high bandwidth. Schuiki et al. [145] proposed a near-memory architecture for training DNNs. This model was developed for accelerating DNN training instead of interference. The training engine, NTX, was used to train the DNNs at scale. They explored the RISC-V cores and NTX coprocessor by reducing the overhead on the main processor by seven times. The NTX combined with the RISC-V processor core offers a shared memory space with single-cycle access on a 128-kB tightly coupled data memory. The architecture employs a hybrid memory cube as the memory module for training the DNNs in data centers.

In [146], a general-purpose vector architecture for migration of ML kernels for near-data processing (NDP) to achieve high speedup with low energy consumption is presented. Their architecture shows a speedup of up to 10x for KNN, 11× for MLP, and 3× for convolution when processing near-data compared to a high-performance ×86 baseline. The work also includes an NDP intrinsics library that supports validating NDP architectures based on large vectors. A machine-learning framework is proposed in [147] to effectively predict the suitable NSP system (among an HBM-based (High Bandwidth Memory) NDP system, an HMC-based (Hybrid Memory Cube) NDP system, and a conventional DDR4-based system) for a given application based on the rankings in performance for a given workload. Kaplan et al. worked on K-means and KNN algorithm evaluation for processing in-storage acceleration of ML (PRINS) [148], a system employing resistive content addressable memory (ReCAM). This architecture functions both as a storage and a massively parallel associative processor. This design works better than the von Neumann architecture model in managing the bottleneck between the storage and main memory. These algorithms outperformed CPU, GPU, and field-programmable gate array (FPGA) in fetching time-accessing data from the main memory. The ReCAM is more efficient than traditional CAMs as it implements line-by-line execution of the truth table of the expression. PRINS enhances the power efficiency and performance compared to other hardware for both K-means and KNN evaluation.

A survey on the architectural aspects, dimensions, challenges, and limitations of In-memory computing processing-in-memory (CIM) is presented in [149]. A robust and area-efficient CIM approach with 6T foundry bit-cells that has improved dynamic voltage range for dot product computations, withstanding bit-cell $V_t$ variations, and eliminating any read disturb issues is proposed in [150]. Recent state-of-the-art works on CIM chips are presented in [151]. As per their research, the SRAM-based CIM solution can be a potential choice for AI processors than NVM-based (non-volatile memory) CIMs. NVM-based CIMs or mem- ristive devices include resistive random-access memory (RRAM), magneto-resistance RAM (MRAM), and phase- change memory (PCM) [152, 153]. A survey on the mem- ristive simulation frameworks, their comparisons, and future modeling is highlighted in [154].

In the past, Cheng et al. [155] introduced the training-in-memory architecture for the memristor-based DNN named TIME. It reduced the computation time of the regular training systems. This architecture supports not only interference but also backpropagation and update during the training of the NNs. It is based on metal-oxide resistive random access memory, which enhances performance and efficiency. The main module is divided into three subarrays: full-function, buffer, and memory. The full-function subarray manages the memory and training operations such as interference, backpropagation, and update. The memory subarray manages data storage, and the buffer subarray holds the intermediate data for the full-function subarray. This architecture improves energy efficiency in deep reinforcement learning and supervised learning.

A thorough survey on hardware accelerators is outside the scope of this paper. Interested readers can refer to [156], a review of accelerators and similar works [157, 158, 159]. However, we could provide the overview of different design aspects at the architecture level to speed up the ML computations

### 6.2. Systolic Arrays

A systolic array is a subset of the data-flow architecture comprising several identical cells, with each cell locally connected to its nearest neighbor. A wavefront of computation is propagated in the array with a throughput proportional to the I/O bandwidth. Systolic arrays are fine-grained and highly concurrent architectures. The progress of IoT-based smart applications has exponentially increased the demand for deep learning algorithms and, in turn, systolic array-based architectures.

In these lines, an automatic design space exploration framework for CNN-based systolic array architecture implementations on an FPGA under high resource utilization and at higher speeds was proposed in [160, 161]. They utilize analytical models to provide in-depth resource estimation and performance analysis. However, systolic array implementations on FPGAs are affected much by the sparsity problem of deep neural networks. Researchers earlier worked towards this problem. An approach of packing sparse convolutional neural networks into a denser format for efficient implementations using systolic arrays is proposed in [162]. However, these designs create irregular sparse models that fail to exploit the data-reuse rate feature of the systolic array. Structured pruning was introduced in [163, 164] to overcome the problem associated with the data-reuse rate that produces DNNs compatible with the synchronous and rhythmic flow of data from memory to the systolic arrays.

Further, [165] propose Eridanus, an approach for structural pruning the zero-values in sparse DNN models before implementing them on systolic arrays. The approach examines the correlation among all the filters to extract the locally-dense blocks, the widths of which match the width of the target systolic array, thus reducing the sparsity problem. Similarly, optimization for the systolic array architecture of deep learning accelerators for sparse CNN models on FPGA platforms is necessary as the zeros in the filter matrix of CNN occupy the computation units resulting in sub-optimal efficiency. A sparse matrix packing method with bit-map representation that condenses sparse filters to reduce the computation required for systolic array accelerators is proposed in [166].

Many systolic array architectural modifications were proposed in the literature addressing specific applications. In [167], an MLP training accelerator as a systolic array on Xilinx U50 Alveo FPGA card is proposed to address the attack detection on a massive amount of traffic logs in network intrusion detection in a short time. The processing speed per power consumption was 11.5 times better than the CPU and 21.4 times better than the GPU. An approximate systolic array architecture combines timing error prediction and approximate computing to relax the timing constraints of MACs [168]. The proposed array on CIFAR-10 image classification could obtain a 36% energy reduction with only a 1% accuracy loss. A reconfigurable systolic ring architecture to reduce on-chip memory requirement and power consumption [169].

Matrix multiplication is one of the primary computations in most computing architectures. [170] proposes a novel systolic array based on factoring and radix-8 multipliers to significantly reduce the area, delay, and power from the conventional radix-4 design providing the same functionality. FusedGCN [171], a systolic architecture that computes the triple matrix multiplication to accelerate graph convolutions. It supports compressed sparse representations and tiled computations without losing the regularity of a systolic architecture. Recently, a hybrid accumulator factored systolic array based on partial factoring of carry propagate adder is proposed [172] with a significant improvement in area, delay, and power.

The functional safety of the accelerators is another critical concern. Faults manifested due to manufacturing defects in the data paths of GPU/TPU accelerated DNNs on systolic arrays may lead to a functional safety violation. An extensive functional safety assessment of a DNN accelerator exposed to faults in the data path is presented in [173].

From the directions of the state-of-the-art works, it demands systolic array architectures that are more flexible, with more data-flow strategies and multiple data transmission modes in the future to handle the increasing depths of deep neural networks.

## 7. AI at the SOC

Artificial intelligence, more specifically deep learning, is feasible in most hardware applications due to the advancements in computing and semiconductor fields. Many attempts have been made to replicate the human brain in next-generation applications, often referred to as neuromorphic computing. Several critical modifications are made to the SoC architectures to incorporate deep-learning capabilities. These design modifications impact general-purpose SoC designs and specialized systems that include specialized processing technologies with heterogeneous and massive parallel matrix computations, innovative memory architectures, and high-speed data connectivity.

AI-SoC models must be compressed to ensure their operation at constrained memory architectures in mobile, communications, automobile, and IoT edge applications. The model compression is performed through controlled pruning without compromising accuracy. However, power, latency, and other areas could be trade-offs. Therefore, the architectural modifications are to be carefully chosen with the combined efforts on memory and datapath subsystems.

FPGA (Field Programmable Gate Array) is one of the widespread and commercially available programmable logic devices to accelerate the computing power of AI on hardware [138, 165]. FPGA became a robust device for hardware accelerators because of its low cost, high energy efficiency, reusability, and flexibility. ASIC (Application Specific Integrated Circuits) are at their best for implementing specialized applications.

NNs are biologically inspired and perform parallel computations. Digital units such as DSP models, floating-point units, ALUs, and high-speed multipliers can be effectively implemented using NN techniques. The fundamental advantage of NNs for digital applications is that high-speed circuits can be realized efficiently because of the almost constant operation time, regardless of the increasing number of bits in the circuit. Exploiting the parallelism in NN computations also provides a balance between using internal and off-chip memory.

Many ML and deep learning applications were reported in the past for the performance evaluation of SoCs. Joseph et al. [174] developed empirical models for processors using LR to characterize the relationship between processor response and micro-architectural parameters. Lee et al. [175], Yun et.al[176] proposed power estimation models established via regression analysis for accurate performance prediction and power of microprocessor applications in the micro-architectural design space. The model proposed in [175] reduces the simulation cost with increased profiling efficiency and improved performance by effectively assessing and modeling the sensitivity according to the number of samples simulated for the model formulation and finding fewer than 4000 sufficient samples from a design space of approximately 22 billion points. Depending on the application, 50% - 90% of predictions achieve error rates of < 10%. The maximum outlier error percent reported is approximately 20% - 33%. Wherein hierarchical Clustering is employed in [176] to determine the best predictors among the ten considered events. The proposed model shows an average estimation error of approximately 4% between the actual and estimated power consumptions when applied to an Intel-XScale-architecture-based PXA320 mobile processor.

An investigation and comparative analysis on the application of Machine Learning algorithms for logic synthesis of incompletely-specified functions is presented in [177]. Periodic performance monitoring of SoCs is essential for high-speed and energy-efficient computing systems. However, performance monitoring is dependent on the accu- rate sampling of critical paths. These critical paths dras- tically vary with PVT conditions, particularly at advanced nodes. Addressing this issue, Wang et. Al [178] proposes a machine-learning-based SoC real-time performance monitoring methodology incorporating physical parasitic characteristics and PVT variations with unknown critical paths.

Several SoC architectures were reported targeting specific applications. MLSoC for multimedia content analysis (implemented in TSMC 90-nm CMOS technology) [179]. Jokic et al. [180] presents a complete end-to-end dual-engine SOC for face analysis that achieves >2X improvement in energy efficiency compared to the state-of-art systems. The efficiency comes with the hierarchical implementation of the Binary Decision Tree in the first level and more power-hungry CNN in the next level, which can be triggered when needed. Machine efficiency monitoring is significant to achieve high productivity, failure, and cost reduction. An SoC-based tool wear monitoring system with a combination of signal processing, deep learning, and decision making is proposed in [181]. The sensor fusion data collected from the three-axial accelerometer and MEMS microphone, combined with the measurement of tool flank wear at different scenarios using a camera, is fed to a CNN to detect any machining variation. Extreme learning machines (ELMs) are NN architectures to increase computational efficiency and performance for large data processing [182]. A low-cost real-time neuromorphic hardware system of spiking Extreme Learning Machine (ELM) with on-chip triplet-based reward-modulated spike-timing-dependent plasticity (R-STDP) learning capability is proposed in [183].

A thorough timing analysis of an SoC is also essential to meet the design specifications. In [184], ensemble learning-based timing analysis in an SoC physical design was performed. Ensemble learning is a combination of multiple machine learning models to improve the performance of the base learners. Many floor plan files with different parameter settings, followed by slack time from Synopsys IC Compiler tool as the label, were used for training supervised learning algorithms. The idea was to feedback on the prediction results at an early stage to the physical design flow to modify the improper floorplan. Bigram-based multi-voltage aware timing path slack divergence prediction [185] utilizes the classification and regression tree (CART) approach. Experimental results show an accuracy of 95 to 97% in predicting cell delays and endpoint timing slack.

CAD tools capable of delivering industrial-quality chip designs must be tuned for optimal PPA (performance, power, area). A holistic approach that involves online and offline machine learning approaches working together for industrial design flow tuning is proposed in [186]. The work highlights SynTunSys (STS), an online system that optimizes designs and generates data for a recommender system that performs offline training and recommendation. The work also proposes adaptable Online & offline Systems for the future that dynamically adapts to the trials originating from the online-learning algorithm and the recommender system in the due lifespan of the system across various STS iterations. Addressing the challenges in meeting timing constraints in modern ICs, Ajirlou et al. [187] proposed an additional ML pipeline stage in the baseline pipelined RISC processor to classify instructions into propagation delay classes and enhance temporal resource utilization. The critical challenges in deploying ML-based SoC design for real design flows are presented in [188]. The work highlights the challenges due to limited data, insufficient open-source benchmarks and datasets, EDA tool-based data generation, and Synthetic data generation.

AI-SoC architectures are at the beginning of their capabilities with tightly coupled processors and memory architectures. There is a long way to reach their full capacity mimicking the human brain in edge applications.

## 8. AI in Physical design

VLSI Physical design has numerous combinatorial problems that require many iterations to converge. Semiconductor technology scaling has increased the complexity of these design problems with complex design rules and design for manufacturing (DFM) constraints, making it challenging to achieve optimal solutions [189]. Traditionally, these issues/violations are detected and fixed manually. However, the traditional manual approach to design closure at advanced nodes is striving hard to meet the market windows. In addition to that, the design quality and manufacturing process in the later stages of the design flow becomes extremely sensitive to the changes in the early stages, in turn increasing the turnaround time and retarding the design closure. Thus, the early-stage prediction of valid designs is critical, particularly at the current technology nodes. Machine learning and pattern-matching techniques provide reasonably good abstraction and quality of results at several stages of physical design. They act as a bridge to connect each step and provide valuable feedback to achieve early design closure.

Broadly, physical design can be divided into four stages: partitioning, floor planning, placement & clock tree synthesis, and routing (Fig. 7). We review AI/ML approaches proposed by the researchers in these stages through the following subsections.

### 8.1. AI for Partitioning, Floor planning and Placement

Partitioning is one of the dominant areas of VLSI physical design. The main objective of partitioning is to divide the complex circuit into sub-blocks, design them individually, and then assemble them separately to reduce the design complexity. Floor planning and placement are the other critical stages in the design flow for design quality and design closure. Floor planning maps the logic description from partitioning and the physical description to minimize chip area and delay. The floor planning goals are arranging the chip's sub-blocks and deciding the type and location of I/O pads, power pads, and power and clock distributions. Placement determines the physical locations of logic gates (cells) in the circuit layout, and its solution largely impacts the subsequent routing and post-routing closure. Global placement, legalization, and detailed placement are the three stages of placement. The global placement provides the rough locations of standard cells, and legalization removes any design rule violations and overlaps based on the global placement solution. Detailed placement incrementally improves the overall placement quality [190].

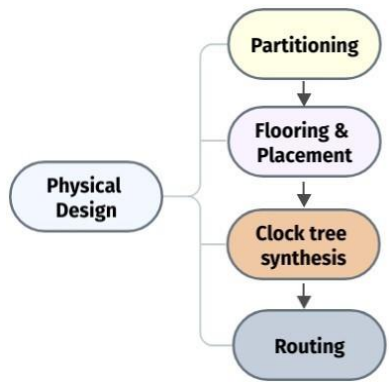

**Figure 7:** Physical Design Flow

Chip floor planning is modeled as a reinforcement learning problem in [191]. An edge-based graph convolutional neural network architecture capable of learning rich and transferable chip representations is modeled out of RL. The method was used to design the next generation of Google's artificial intelligence accelerators and has shown the potential to save thousands of hours of human effort for each new generation. A machine learning-based methodology to predict post P&R (place and route) slack of SRAMs at the floor planning stage, given only a netlist, constraints, and floor plan context tested on 28nm foundry FDSOI technology shows a worst-case error of 224ps [192]. Cheng et al. propose [193] regression methodology to quickly evaluate routing congestion and half-perimeter wire length in each macro placement during floor planning. They explored solutions using different regression techniques – LR, DTR (decision tree regressor), booster DTR, NN, and Poisson regression. Among these, DTR showed a better performance. A multi-chip module (MCM) has many small chips integrated into a package and joined by interconnects [194]. Multi-chip partitioning is harder due to sparse search space. An RL solution for partitioning ML models in MCM is presented in [195].

Moving to placement, high regularity of data paths is essential for compact layout design during placement. However, the data paths are frequently mixed with other circuits, such as random logic. For designs with many embedded data paths, it is crucial to extract and place them appropriately for high-quality placement. Existing analytical placement techniques handle them sub-optimally [196]. However, modern placers fail to handle data paths effectively due to technological constraints. ML plays a crucial role in such scenarios. Ward et al. [197] proposed PADE to demonstrate the capability of automatic datapath extraction for large-scale designs mixed with random and datapath circuits. The effective features are extracted by analyzing the global placement netlist to predict the direction of the datapath. PADE employs a combination of SVM and NN for cluster classification and evaluation. Experimental results on hybrid benchmarks showed promising improvements in half-perimeter and Steiner tree wire lengths. Wang et al. present a connection vector-based and learning-based data path logic extraction strategies [198]. SVM and CNN are employed for machine learning based extraction. Results on MISPD 2011 data path benchmarks show that both the strategies equally perform in classifying data path and non-data path parts.

Chip placement is one of the chip design cycle's most time-consuming and complex stages. AI will provide the necessary means to shorten the chip design cycle, ultimately forming a symbiotic relationship between the hardware and AI, each promoting the advancement of the other. To reduce the time required by the chip placement, Mirhoseini et al. proposed an approach that can learn from past experiences and improve over time [199]. The authors posed placement as an RL problem and trained an agent to place the nodes of a chip netlist onto a chip canvas such that the final PPA is optimized while adhering to the constraints imposed by the placement density and routing congestion. The RL agent (policy network) sequentially places the macros, and once all macros are placed, a force-directed method produces a rough placement of the standard cells. This RL agent becomes faster and better at chip placement as it gains experience on numerous chip netlists. The results ensured that the proposed approach generates placements in under 6 hours, whereas the strongest baselines require human experts in the loop, and the overall process may take several weeks. In [200], quantum machine learning techniques are proposed for faster and optimal solutions with low-error rates to VLSI placement problems. A complete placement was achieved using the variational quantum Eigen solver (VQE) [201] approach, tested on two circuits: a toy circuit (comprising eight gates) and another circuit called "Apte," taken from the MCNC benchmark suite [83]. Research on GPU acceleration for placement and timing analysis achieved 500x speedup for static timing analysis on a million-gate design harnessing the power of machine learning techniques with heterogeneous parallelism [202].

Placement and routing are two highly dependent physical design stages. Tight cooperation between them is highly recommendable for optimized chip layout. Traditional placement algorithms that estimate routability using pin delay or through wirelength models can never meet their objectives due to increased manufacturing constraints, and complex standard cell layouts [203]. A deep-learning model (CNN based) to estimate the routability of a placement to quickly analyze the degree of routing difficulty to be encountered by the detailed router is presented in [204]. In [205], a CNN-based RL model is proposed for detailed Placement, keeping optimal routability for current Placement. A generalized placement optimization framework to meet the post-layout PPA metrics with a small runtime overhead is proposed in [206]. Given an initial placement, unsupervised learn- ing discovers the critical cell clusters for post-route PPA improvements from timing, power, and congestion analysis. A directed-placement optimization followed them. The approach is validated on industrial benchmarks in a 5nm technology node.

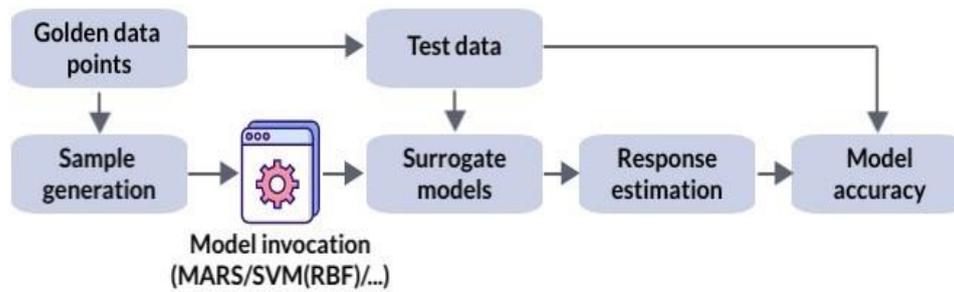

**Figure 8:** Meta Modeling Flow

A machine-learning model for predicting the sensitivity of minimum valid block-level area of various physical layout factors that provides 100x speedup compared to conventional design technology co-optimization (DTCO) and system technology co-optimization (STCO) approaches is proposed in [207]. This research suggests bootstrap aggregation and gradient boosting techniques for block-level area sensitivity prediction from their experiments across various ML algorithms. Further, [208] quotes MAGICAL (Fully automated analog layout from netlists to GDSII, includ- ing automatic layout constraint generation, placement, and routing), an open-source VLSI placement engine. Magical 1.0 is open-source. [209] presents automated floor planning by exploration with different floor plan alternatives and placement styles.

RL is being proposed as the best solution for the physical design of an IC as it does not depend on any external data or prior knowledge for training and could produce unusual solutions based on the design space exploration by the agent. Some RL approaches for placement optimizations [210, 211].

### 8.2. AI for Clock Tree Synthesis(CTS)

Clock tree synthesis is one of the crucial steps in the VLSI physical design. It is used to reduce clock skew and insertion delay. As the clock network contributes a large percentage of the overall power in the final full-chip design, it is vital to have an optimized clock tree that prevents serious design problems, including excessive power consumption, high routing congestion (caused when extra shielding techniques are used), and protracted time closure. With the downscaling of devices, the run time and complexity of existing EDA tools for accomplishing CTS have increased. Highly efficient clock trees that optimize key-desired parameters, such as the clock power, skew, and clock wire length, are required. It is a very time-consuming process involv- ing searching for parameters in a wide range of candidate parameters. Several ML algorithms have been proposed to automate the prediction of clock-network metrics.

Data mining tools such as the cubist data mining tool [212] are used to achieve skew and insertion delay efficiently. In [213], statistical learning and meta-modeling methods (including surrogate models) were employed to predict essential parameters, such as the clock power and clock wire length, as shown in Fig. 8. In [214], the authors implement a hierarchical hybrid surrogate model for CTS prediction, mitigating parameter multi-collinearity challenges in relatively high dimensions. They tackle the high-dimensionality problem by dividing the architectural and floor planning parameters into two groups – one with low multi-collinearity and the other with parameters that exhibit large linear dependence. Later the models from these groups are combined through least-squares regression (LSQR). [215] presents an ANN-based transient clock power estimation that can be applied to pre-CTS netlists.

Ray et al. [216] employ ML-based parameter tuning in multi-source CTS to build a high-performance clock network with a quick turnaround time. GAN-CTS, a conditional GAN framework for CTS outcome prediction and optimization, outperforms commercial auto-generated clock tree tools in terms of clock power, skew, and wire length (target CTS metrics) [217]. Design features are directly extracted from placement layout images to perform practical CTS outcome predictions. The framework also employs RL to supervise the generator of GAN toward clock tree optimization. A modified GAN-CTS [218] employs a multitask learning technique to simultaneously predict the target CTS metrics using multi-output deep NN. It achieves higher accuracy in a shorter training time compared to the meta-learning approach [217]. An RL-based solution reduces over 40% of the peak current of a design at the CTS stage compared to the heuristic CTS solutions utilized by physical design EDA tools [219].

## 8.3. AI for Routing

Routing lays physical connections to the circuit blocks and pins assigned during the placement as per logical connectivity and design rules. Global routing (GR) and detailed routing are two routing stages. Global routing partitions the routing region into tiles and decides tile-to-tile paths for all nets while attempting to optimize specific objectives such as minimum wire length and timing budget. The actual geometric layout of each net within the assigned routing regions is carried out in the detailed routing stage [220].

Routing congestion [221] is the major bottleneck in the GR stage which is caused when an overflow of net assignment occurs in a region. Another area for improvement in routing is DRVs (detailed routing violations). The heuristic and probabilistic approaches employed in traditional GR solutions [222, 223] suffer from scalability limitations associated with advanced nodes. Early prediction of routing requirements enables the design engineers to create high-quality layouts faster. Some research efforts were made to address these challenges using machine-learning-based approaches.

MARS (multivariate adaptive regression splines), a non-parametric flexible regression modeling for high-dimensional data, is used for modeling routing congestion in [224]. Qi et al. [225] also utilize MARS to construct a routing congestion model that directly estimates detailed routing congestion through a mapping function that maps global routes and layout data to detailed routing congestion. Router-friendly placement solutions can be obtained from congestion estimators. SVM for classifying BEOL (back end of line) stack-specific placements routability based on the DRVs/DRCs (design rule check) from P & R tools at the post-route stage achieved significant improvements than employing only congestion maps [226]. Xie et al. propose RouteNet [227] to evaluate the overall routability of cell placement solutions without global routing or predict the locations of DRC (Design Rule Checking) hotspots. RouteNet is built over CNNs and shows 50% higher accuracy than GR. An ML approach predicts any short violations before detailed routing with placement and global routing congestion information and sends it as feedback to the placement system for improvement [228, 229, 230].

Figure 9 shows the general procedure of ML-based routing shorts prediction. In the routing step, each circuit under training is routed using a detailed routing tool, and the locations of any identified shorts are collected. In feature extraction, the circuit area is divided into small tiles, and occurrences of shorts are investigated inside these tiles. Each tile is described by a feature vector (appropriate features that contribute to routing violations) and is considered as an instance. An instance belongs to the negative (N) class and is labeled with the target value of 0, i.e., there is no short in its tile area. An instance belongs to the positive (P) class and is labeled with the target value of 1 if any short violation is observed in its tile area after detailed routing. The collected data are fed to a supervised-learning algorithm.

Zhang et al. propose a density and pins peaks-based fast neural network algorithm in NTHU-Route 2.0 [231] to predict congestion map [232]. A GAN-based congestion estimator can produce congestion heatmaps from placement, and netlist information [233]. A deep learning framework for predicting the shorts violations by extracting useful features after placement and analyzing them drastically decreases the prediction time and requirement of a global router [234]. Interestingly, the framework considers a short prediction problem as a binary classification problem with imbalanced data. The results show that the model is 14x faster than NCTU-GR [235] for smaller designs and up to 96x faster for larger designs. A CNN-based GR congestion estimation algorithm [236] that utilizes the 3D congestion information similar to [237] showed an incremental improvement in the number of overflows, wire length, and vias. The congestion heatmaps and placement information extracted from hyper-images of the design feature extraction algorithm act as inputs to the congestion model. Goswami et al. proposed a regression-based routing congestion prediction problem for FPGAs [238]. The paper reports important features through thorough feature engineering for modeling the regression algorithms - RF, MLP, LR, and MARS. On average, the proposed methodology is 25 to 50 times faster than Xilinx Vivado-based routing calculation tool, which reports actual congestion after detailed routing.

One solution for reducing the overall time of physical design is to predict the circuit performance after physical design. Li and Franzon [239] proposed an ML approach using surrogate modeling (SUMO). They employed surrogate models to predict the results after the GR step. In the first stage, SUMO generates models for each output to predict the GR results in the future. In the second stage, after analyzing the linear relationships among thousands of GR results and detailed routing results, these results were set as inputs and outputs in ML models. These trained ML models precisely predict the after-detailed routing results using the GR results.

NNs and decision trees are the most used ML models for this problem. A machine learning-based pre-routing timing prediction approach [240] shows a closer match with post-routing analysis from Synopsys PrimeTime. RF performed well in their analysis compared to lasso and NN techniques.

Substrate routing automation framework through supervised learning algorithms is proposed in [241]. It combines manual and automated results as training data to a CNN for improved design cycle and performance. A GNN-based congestion estimation approach that can predict the detail routed lower metal layer congestion values from a technology-specific gate-level netlist for every cell in a design is proposed in [242]. The training dataset is built from the detail-routed congestion maps by dividing them into discrete grids and assigning the congestion value of each grid as the target value. Another GNN-based routing short violations prediction at the placement stage is proposed in [243]. The information is fed back to the placement system, and a new placement result is generated with reduced DRC violations. GraphSAGE (Graph sample and aggregate) is applied effectively to combine the adjacency matrix with the features of each tile. [244] presents a survey of the recent development of machine learning-based routing algorithms. XGBoost is employed to predict post-detailed routing timing at the post-GR stage in [245]. When employed for post-GR optimization, it improves the circuit performance.

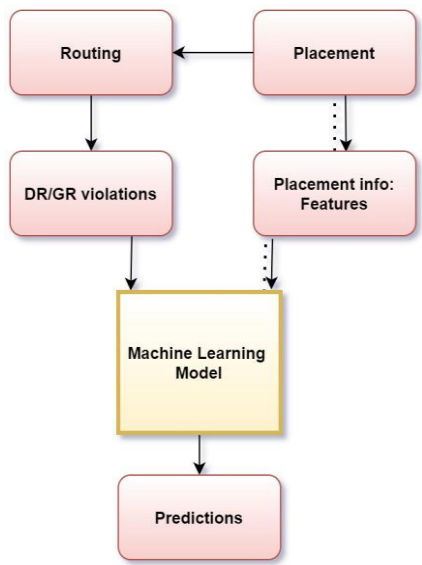

**Figure 9:** Typical flow of ML-based routing techniques

Supervised learning, NNs in particular, and RL-based solutions reported in the literature produced many valuable feedbacks and solutions for different complex modeling tasks at various physical design stages.

After placement and routing, the layout is generated and the design is ready for fabrication.

## 9. AI in Manufacturing

Numerous processes are involved in manufacturing an IC, including wafer preparation, epitaxy, oxidation, diffusion, ion implantation, lithography, etching, and metallization [246]. All the steps are performed in highly sophisticated fabrication units with constant human supervision. The fabricated ICs are packaged in special packages to protect them from external/environmental damage.

### 9.1. AI for Lithography

Most chip-manufacturing processes are complex chemical processes, except for the lithography process. Lithography transforms layout data into geometric patterns as masks and from masks to the resist material on the semiconductor. After the physical design, lithography is a crucial step in chip manufacturing. Masks identify spaces on the wafer where certain materials need to be deposited, diffused, or removed. The fabrication process involves several dozen steps of deposition and diffusion based on the circuit's complexity. During each step, one mask is used. The exposure parameters required to achieve accurate pattern transfer from the mask to the photosensitive layer primarily depend on the wavelength of the radiation source and the dose required to achieve the desired change in the properties of the photoresist. Identifying defects during mask synthesis and verifying each lithography stage before proceeding to the next stage is crucial for yield enhancement but very difficult in the nanometer dimensions, mainly due to the increased random variations in the process. Introduction/improvement of automated procedures at various stages of lithography is necessary to increase the manufacturing yield and reduce the cost and turnaround time. Traditionally, this was a very laborious process; fortunately, the introduction of ML has afforded many opportunities for increasing the processing speed, particularly in mask synthesis and verification [247].

The need for Machine Learning in the lithography process is discussed in [248]. It also highlights various algorithms and their trade-offs used for hotspot detection (HD), optical proximity correction (OPC), sub-resolution assist feature (SRAF), phase shift masks (PSM), and resist modelling. They also propose a Gaussian process to reduce the false positive outcomes of ML algorithms. In detail, we discuss the research on ML-Lithography in the following sub-sections.

#### 9.1.1. At Mask Synthesis

Optical lithography is the most widely used technique in IC manufacturing, where a geometric mask is projected into a photo-resist-coated semiconductor through a photon-based technique. Moore's law has driven features to ever smaller dimensions, and the technology has been scaled down to the limit of light wavelength. Consequently, the printed patterns get distorted due to diffraction, resulting in process defects. Various resolution enhancement techniques (RETs) are employed to improve the performance of photo-lithography. OPC and SRAF insertion are the most used RETs to maximize the process window and ensure accurate patterns on the wafer. However, these enhancement techniques suffer from an extremely long runtime owing to their laborious iterative process. Many state-of-the-art methods use machine learning to identify defective lithographic patterns.

LR was the first ML technique used in OPC. An LR model for predicting the optimum starting point for a traditional-iterative-model-based OPC has been proposed [249]. Using discrete cosine transform coefficients from the lowpass-filtered 2 × 2 um layout patterns as inputs and creating separate models for normal edge, concave corner, and convex corner fragments; they achieved a 32% reduction in the runtime. When Luo [250] proposed a three-layer MLP to generate the optimal mask pattern for OPC, deep learning came into use. Using the steepest descent method to generate the training set, his model drastically reduced computation time.

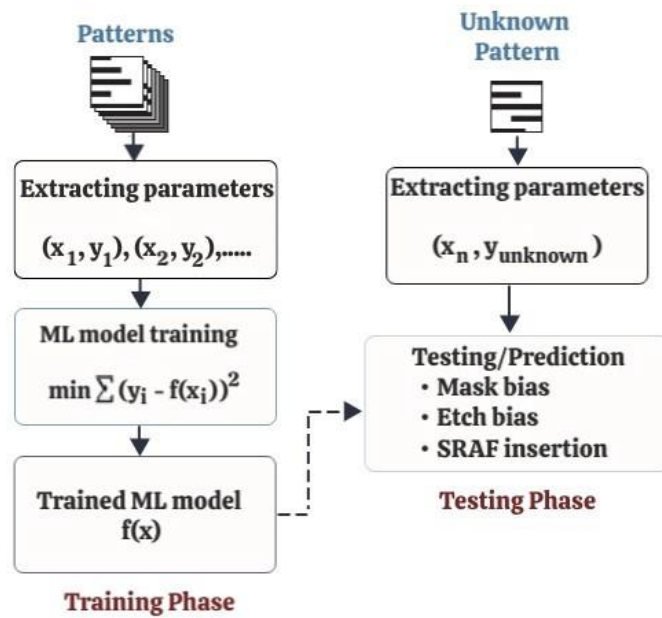

**Figure 10:** Typical procedure of ML-based mask Synthesis Flow

A hierarchical Bayes model (HBM) was proposed in [251], for OPC, along with a new feature-extracting technique known as concentric circle area sampling (CCAS). HBM provides a flexible model that is not constrained by the linearity of the model parameters or the number of samples; this model utilizes a Bayes inference technique to learn the optimal parameters from the given data. All parameters are estimated using the Markov chain MC method [252]. This approach has shown better results than other ML techniques, such as LR and SVMs. Most ML OPCs use local pattern densities or pixel values of rasterized layouts as parameters, which are typically huge numbers. It leads to overfitting and, consequently, reduced accuracy. Choi et al. [253] proposed the usage of basic functions of polar Fourier transform (PFT) as parameters of ML OPC. The PFT signals obtained from the layout are used as input parameters for an MLP whose number of layers and neurons are decided empirically. Experimental results show that this model achieves an 80% reduction in the OPC time and a 35% reduction in error.

ML is also explored in inverse lithography technology (ILT) [254], a popular pixel-based OPC method. ILT treats the OPC as an inverse imaging problem and follows a rigorous approach to determine the mask shapes that produce the desired on-wafer results. Jia and Lam [255] developed a stochastic gradient descent model for mask optimization that showed promising results in robust mask production. Luo et al. [256] proposed an SVM-based layout retargeting method for ILT for fast convergence. A solution to ILT was achieved through a hybrid approach by combining physics-based feature maps [257] with image space information as model inputs to DCNN (deep CNN) [258].

SRAFSs are small rectangular patterns on a mask that assist in printing target patterns; they are not printed even though they are on the mask. The process of SRAF generation is similar to OPC and is computationally expensive. Recently, ML was applied to SRAF generation. Xu et al. [259] demonstrated an SRAF generation technique with supervised-learning data for the first time. In their model, features are extracted using CCAS and compacted to reduce training data size. Logistic regression and SVM models were employed for training and testing. Instead of using binary classification models, the author uses the models as probability maxima. SRAFs are inserted at the grids with the probability maxima. This model shows a drastic speedup in computation with less error. Shim et al. [260] used decision trees and logistic regression for SRAF generation, which showed a 10× improvement in runtime.

Etching and mask synthesis are performed simultaneously. Recently, ML has been used to predict the etch bias (over-etched or under-etched). ANNs [261, 262, 263] have been used to predict the etch proximity correction to compensate for the etch bias, yielding better accuracy than traditional methods.

Although these ML models achieve high accuracy, they require a large amount of data for training. In the field of lithography, where the technology shrinks very rapidly, and old data cannot be used for the new models, data generation is a very laborious task. One of the solutions to this problem is to use transfer learning, [264] which takes the data generated through old technology nodes and information about the evolution of nodes, e.g., from 10 to 7 nm, and uses them for model training. The authors also employ active data selection to use the unlabeled data for training using Clustering. ResNet is used along with these two active learning and transfer learning techniques, yielding high accuracy with very few data samples for training.

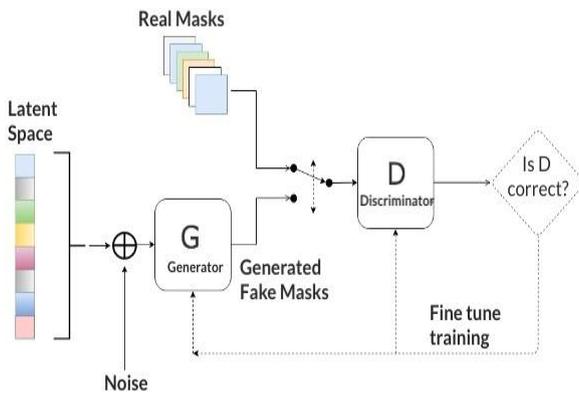

**Figure 11:** Generative Adversarial Networks

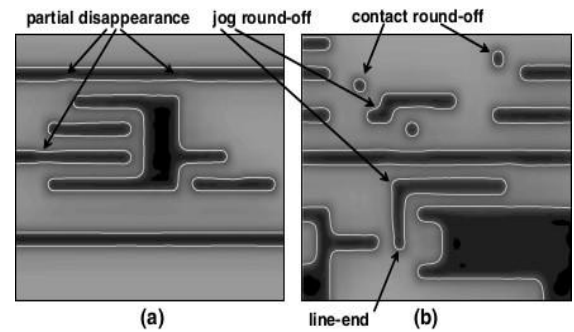

**Figure 12:** Examples of lithography hotspot Patterns

GANs [53] are one of the hottest prospects in deep learning. Figure 11 shows the general design of the primary optimization flow of a generative adversarial network. It contains two networks interacting with each other. The first one is called the "generator" and takes random vectors as input and generates samples as close to the true dataset distribution as possible. The second one is called the "discriminator" and attempts to distinguish the true dataset from the generated samples. At convergence, ideally, the generator is expected to generate samples with the same distribution as the true dataset. This technique was exploited in lithography modeling. GANs were used for OPC where intermediate ILT results initialize the generator; this improves the training process, allowing the network to produce an improved mask [265]. In [266], CGAN was used for the generation of SRAF. Conditional GAN is an extension of GAN, where the generator and discriminator are conditioned on some auxiliary information, such as class labels or data, from other modalities. A new technique for data preparation, i.e., a novel multi-channel heatmap encoding/decoding scheme that maps layouts to images suitable for CGAN training while preserving the layout details, was also proposed here. This model achieves a 14× reduction in computation costs compared to state-of-the-art ML techniques. LithoGAN [267] is an end-to-end lithography modeling approach where the mask pattern is directly mapped to the resist pattern. Here, a CGAN is used to predict the shape of the resist pattern, and a CNN is used to determine the center location of the resist pattern. This technique overcomes the laborious process of building and training a model for each stage, resulting in a reduction in computation time of approximately 190 times compared to other ML techniques.

Different OPC engines work on different design patterns, each of which has advantages and disadvantages. Compared to the model-based OPC, ILTs generally promise good mask printability owing to their relatively large solution space. However, this conclusion only sometimes holds as ILTs need to solve a highly non-convex optimization problem, which is occasionally challenging to converge. GAN yields good results; however, it is difficult to train for some patterns. To overcome these challenges, Yang et al.[268] proposed a heterogeneous OPC flow, where a deterministic ML model decides the appropriate OPC engine for a given pattern, taking advantage of both ILT and model-based OPC with negligible overhead. They designed a classification model with a task-aware loss function to capture the design characteristics better and achieve their objectives. Yang et al. [269] also proposed an active-learning-based layout pattern sampling and HD flow for effective, optimized pattern selection. The experiments show that the proposed flow significantly reduces the lithography simulation overhead with satisfactory detection accuracy.

E-beam lithography is another prominent patterning method to electronically transfer the layouts onto the wafer. Non-uniformities caused by parallel e-beam maskless lithography result in variations within the targets. Scatterometry measures the defects caused by simulated dose variations in patterned multi-beam maskless lithography. An ML-based scatterometry to quantify critical dimension (measured parameter for variation detection) and sensitivity analysis in detecting beam defects is proposed in [270]. A fast in-line EUV resist characterization using scatterometry in conjunction with machine learning algorithms is presented in [271].

*9.1.2. At Mask Verification*

Due to complicated design rules and various RETs such as OPC and SRAF, there may still be many lithographic hotspots that may cause opens, shorts, and reductions in yield (Fig. 12). Therefore, detecting and removing these hotspots are critical for achieving a high yield. Traditionally, pattern-matching techniques are widely used in HD. Hotspot patterns are stored in a predefined library, and given a new testing pattern; a hotspot is detected if it can be matched to the existing patterns. This technique is accurate for already-known hotspot patterns but does not work well for new, unknown patterns. ML-based approaches show better accuracy for both seen and unseen patterns.

Early ML usage in lithography HD included classifiers such as simple NNs (including ANNs) [272, 273], which detect hotspots from given patterns. Clustering algorithms were also extensively used, [274, 275], where a large dataset of hotspots is divided into multiple classes using these algorithms, and pattern-matching techniques are used for the detection of new hotspots. As the detected hotspots in the same class share similar geometric shapes, it is expected that they can be fixed using a standard fixing solution. False alarms are a critical issue in HD in many ML methods. Researchers have been attempting to overcome this challenge. Ding et al. [276] attempted to successfully refine the SVM and ANN classifiers to identify the hotspot patterns more accurately Topological classification is another method where feedback learning is employed to reduce false alarms. Yu et al. [277] classified the already-known hotspots and non-hotspot patterns into clusters according to the topologies in their core regions. Subsequently, they extracted critical features and constructed an SVM kernel with multiple feedback learning from the mispredicted non-hotspots. Combining different ML techniques yields better outcomes most times. Ding et al. [278] proposed a new algorithm that combines ML and pattern matching, and Matsunawa et al. [279] used the AdaBoost classifier; both approaches resulted in a significant reduction in false alarms and outperformed many other ML techniques. Semi-supervised learning [280] is also being used for detection; it leverages both labeled and unlabeled data, thereby reducing the dependence on labeled training data. It is advantageous as obtaining labeled hotspot regions is considerably more difficult. This method combines Classification and Clustering, creating a multitasking network that groups the unlabeled data with labeled data and then uses them for learning. It reduces the pre-processing time and amount of labeled training data required.

CNNs are a widely used NN technique in image processing, classification, etc. HD is very similar to image classification; CNNs [281, 282] have recently been used in this field, yielding better accuracy than other state-of-the-art ML approaches. Pooling layers are one of the building blocks of CNNs. These layers reduce the number of parameters and computation steps in the network by extracting the statistical summary of the local regions of the previous layer, thereby reducing the feature map dimension and drastically lowering the sensitivity of the NN to small changes. However, in the HD process, these layers may ignore small edge displacements and turn them into non-hotspot regions. Yang et al. [283] proposed a pooling-free CNN architecture that overcomes this defect, yielding increased accuracy. Online learning is another method in ML where the model is trained and updated with new data that is fed over time to build the predictor for future data. This method can adapt over time and works well with new models; thus, it can be used in HD. Although CNN has the potential to perform well in HD, hotspot patterns are always minorities in the VLSI mask design as less number of patterns are available for training, resulting in an imbalanced training data set; this results in a model with high false negatives.

Yang et al. [284] attempted to apply minority up-sampling and random-mirror flipping before training the network and achieved better performance than state-of-the-art hotspot detectors. In this pre-processing technique, the training dataset is first augmented with a mirror-flipped, 180°-rotated version of the original layout clips, followed by up-sampling. In this technique, overfitting can be reduced through random mirroring. Zhang et al. [285] built an online learning model with a novel critical feature extraction technique. They constructed an ensemble classifier using smooth boosting and modified Naive Bayes. Their technique outperformed state-of-the-art methods in terms of accuracy and false alarms. Subsequently, they extended this technique to on-line learning, which yielded even better performance. Ye [286] proposed litho-GPA, a Gaussian process assurance (confidence value), provided along with each prediction. The framework also incorporated a set of weak classifiers and active data sampling for learning, reducing the amount of training data and computations required.

Park et al. [287] propose an SVM model trained with lithographic information that detects pinching and bridging hotspots during mask transferring to wafer. They further incorporate domain knowledge of lithographic information in the SVM kernel to accomplish an accurate decision function to classify them into four categories – horizontal bridging (HB), vertical bridging (VB), horizontal pinching (HP), and vertical pinching (VP). A hybrid pattern matching-SVM classifier for HD is presented in [288]. CNN-based HD is proposed in [289, 290]. The framework in [290] also has a transfer learning scheme to reduce the training sample requirement for modeling HD at a more advanced node. A modified DNN by replacing pooling layers with convolution layers for HD is proposed in [291]. They applied hotspot folding, rotating, and mirror-flipping for highly imbalanced datasets to maximize the training samples. Addressing the challenge of an imbalanced dataset in HD, [292] propose a dataset sampling technique based on autoencoders. The autoencoders identify latent data features that can reconstruct the input patterns, which are then grouped using Density-based spatial clustering of applications with noise (DBSCAN). These clustered patterns are sampled to reduce the training set size. An automatic layout generation tool that can synthesize different layout patterns given a set of design rules is proposed in [293]. The tool supports via and unidirectional metal layer generation. It's robustness in HD is tested using state-of-the-art ML models.

SONR (state of nature reduction) [294], a semi-supervised feature vector-based ML tool for lithographic HD at different stages and cross products based on known hotspots, is a fast and effective method to optimize OPC verification flow and improve manufacturing yield. The proposed workflow is available as Mentor's Calibre SONR tool. [295] demonstrates an HD case study based on ADAPT, a framework for the fast migration of machine learning models across different IC technologies. It is an unsupervised Bayesian approach to significantly reduce model cost and provide customized learning with fewer data techniques and labeling strategies. An ML-based color defect detection for after develop inspections in lithography exhibited more sensitivity and specificity in a trial comparison against the reference method [296, 297].

AI/ML in VLSIAutomation in SEM (scanning electron microscope) image pre-processing using dimensionality reduction and feature detection dramatically reduces the computation time of lithography patterning [298]. A framework combining ML models for automatically mining lithographic hotspots from massive SEM images detects hard defects such as bridging and necking and soft defects such as scumming that are hard to detect by manual inspection [299]. However, they propose manual inspection on top of their framework for the final decision on the detected hotspots. The solution proposed could reduce the workload to a large extent compared with the traditional way. Recently, many researchers have been searching for efficient solutions beyond ML [300, 301]. A circuit-based hybrid quantum-classical machine learning using variational quantum layers for lithography HD from SEM images is proposed in [300]. The hybrid approach adds quantum circuits to the conventional CNN for enhanced performance. Quantum computing simulation has been performed with CuQuantum, an Nvidia software development kit with optimized libraries and tools for accelerating quantum computing workflows. Virtual metrology model using CNNs to predict the overlay errors of the photo-lithography process [301].

Layout patterns play an essential role as resources for flows of various DFM that we have already discussed. However, VLSI layout pattern libraries are not readily available due to the long and iterative technology life cycle, which can slow down the technology node development. However, significant effort has been devoted to enlarging existing libraries by exploiting existing patterns, including flipping, rotating, and using a random generator. These methods are coupled with complex manuals for guidance and hardly increase the layout diversity owing to their deterministic strategy. To address these problems, Zhang et al. [302] proposed a pattern generation and legalization framework comprising two learning-based modules for pattern topology generation and design rule legalization. In the generation stage, a variational convolutional autoencoder (VCAE) [303] is designed to efficiently generate realistic pattern topologies via Gaussian perturbation. For the legalization stage, a CGAN [304] model is used to transform the generated samples from blurry patterns to smooth ones, significantly reducing the DRC violation risks. Based on an adversarial autoencoder, a pattern style detection tool is designed to examine the pattern styles and filter out unrealistic generated patterns. A novel confidence-aware deep learning model for post-fabrication wafer map defect is proposed in [305]. The experiment results on industrial wafer datasets demonstrate superior accuracy compared to the traditional approach. The paper also discusses the scope of DL-based approaches for manufacturing and yield in the near future.

Evidently, ML is no longer a novelty in chip fabrication. Chip manufacturers will continue to leverage the technology as it matures. ML provides solutions to many problems in lithography. However, unlike areas such as image processing, where a large amount of data is available, it is difficult and expensive to obtain enough data in VLSI design for training robust and accurate models. Therefore, developing techniques for improving modeling accuracy with a relaxed demand for big data is critical to promote the widespread adoption of ML.

## 9.2. Reliability Analysis

Over the last few decades, shrinking CMOS geometries have increased manufacturing defect levels and on-chip fault rates. Increased fault rates have considerably impacted the performance and reliability of circuits. The multiplied fault rates have necessitated an accurate and robust reliability analysis. The fundamental reliability analysis evaluates logic circuit errors due to hot-carrier insertion, electro-migration, NBTI (negative-bias temperature instability), and electrostatic ejection. Reliability engineers focus on correcting the functionality and enhancing the circuit's lifetime.

Precise reliability analysis involves numerous mathematical equations. However, mathematical equations fall apart due to the complexity involved in the reliability estimation of large circuits with millions of gates. Reliability engineers have worked on MC simulations, which are rigorous and time-consuming. Therefore, the evolution of ML has aided engineers in developing exhaustive and rapid reliability analysis algorithms. Patel et al. [306] and Krishnaswamy et al. [307] developed probabilistic transfer matrices, which perform simultaneous analyses over all the possible I/O combinations. The major limitation of the method is the large memory requirement for the matrices. Choudhury et al. [308] recommended algorithms for reliability analysis based on a single pass, observability, and a max-k gate. The methods are precise for PVT and aging-related degradation. Beg et al. [309] presented an NN-based nanocircuit reliability estimation method as an alternative to traditional mathematical methods. This method is time efficient for the analysis of the circuits.

Circuit aging is one of the essential concerns in the nanometer regime in designing future reliable ICs. Several operating conditions, such as temperature, voltage bias, and process parameters, influence the performance degradation of the IC. NBTI is a significant phenomenon occurring at present and future technology nodes and contributes significantly to the performance degradation of an IC due to aging. It shifts the threshold voltage during the lifetime, degrades the device drive current, and degrades the device performance. It is necessary to evaluate the impact of NBTI on the performance of a circuit under stress early in the design phase to incorporate appropriate design solutions. Many researchers have contributed to the NBTI estimation early in the design phase through ML algorithms.

Karmi et al. [310] proposed an aging prognosis approach based on nonlinear regression models that map circuit operating conditions to critical path delay estimation. The approach also considered the effects due to process variations. The experiments showed that the impact of IC aging on critical path delays could be accurately estimated through nonlinear regression models. Such modeling facilitates the

D Amuru et al.: *Preprint submitted to Elsevier*Page 23 of 41

implementation of preventive actions before the circuit experiences aging-related malfunctions. A gate-level timing prediction under dynamic on-chip variations is proposed in [311]. The high-dimensional features added to the statistical timing analysis for modeling the NBTI increase with increasing circuit complexity. The proposed learning-based approach efficiently captures these high-dimensional correlations and estimates the NBTI-induced delay degradation, with a maximum absolute error of 4% across all the designs. SVR and random forest models were applied to the timing estimation. Analysis and estimation of the impact of NBTI-induced variations at multi-gate transistors in digital circuits are becoming highly challenging [312]. A quick and accurate estimation of process variation impact and device aging on the delay of any path within a circuit is possible through GNNs [313].

Electro-migration is another concern successfully addressed by various AI strategies. A new data-driven learning-based approach for fast 2D analysis of electric potential and electric fields based on DNNs is proposed in [314]. As an extension, Lamichhane et al. [315] proposed an image-generative learning framework for electrostatic analysis for VLSI dielectric aging estimation. It speeds up the analysis compared to the conventional numerical method, COSMOL. Compared to the similar CNN-based method, the proposed GAN-based approach gives 1.54x more speedup with around similar accuracy.

Reliability analysis and failure prediction of 3D ICs has gained considerable attention over the past few years. A study on the 3D X-ray tomographic images combined with AI deep learning based on a CNN for non-destructive analysis of solder interconnects demonstrates an accuracy of 89.9% in predicting the interconnect operational faults of solder joints of 3D ICs [316]. Adaptive lifetime prediction techniques (ADLPT) that minimize redundant prediction operations in 3D NAND flash memories by exploiting reliability variation are presented in [317].

Kundu et al. [318] confer the reliability issues of different AI/ML hardware. The paper explores and analyzes the impact of DRAM faults on the performance of the DNN accelerator by implementing MLP on MNIST datasets. Further, they discussed the impact of the circuit and transistor-level hazards such as PVT variations, runtime power supply voltage noise and droop, circuit aging, and radiation-induced soft errors on AI/ML accelerator performance. The accuracy impact on MAC units due to these hazards has been estimated. The paper also highlights the reliability issues of neuromorphic hardware and proposes RENEU, a reliability-oriented approach to map machine learning applications to it.

The ever-growing circuit complexity is also raising concerns about hardware security. ML can aid in detecting hardware attacks and could take necessary counter-attacks with suitable design [319]. Hardware assurance and verification in manufactured ICs are also important to identify hardware Trojans. Manual verification to identify such security threats is becoming challenging at the present scale of circuit design. Addressing these issues, [320] proposed CNN-based arithmetic circuit classification, taking the image generated from a circuit's conjunctive normal form description. However, the structural information of circuits is difficult to capture in the CNN framework. Resolving it, [321] proposes a GNN framework for ASIC circuit netlist recognition, which classifies circuits according to their structural similarity. Case studies on four designs of adder circuits exhibit 98.3% accuracy. Several environmental, performance and process-related embedded instruments (EI) are present in an SoC with a JTAG interface. The EI data is systematically collected over time and analyzed using PCA (principal component analysis) and a power-law-based degradation model to predict the remaining valid lifetime of an SoC [322]. Liakos et al. [323] proposed hardware trojan learning analysis (ATLAS) that identifies hardware trojan-infected circuits using a gradient boosting model on data from the gate-level netlists phase. The feature extraction was based on the area and power analysis from Synopsis Design Compiler NXT industrial tool. ATLAS model was trained and tested on all circuits available in the Trust-HUB benchmark suite. The experimental results show that the classification performance is better than the existing models. In [324], GNNs are proposed for reverse engineering of gate-level netlists without manual intervention or post-processing. The experimental results on EPFL benchmarks [325], the ISCAS-85 benchmarks, and the 74X series benchmark show an average accuracy of 98.82% in terms of mapping individual gates to modules.

Numerical methods and MC simulations, which reliability engineers widely use, have memory and timing constraint bottlenecks. The NNs and Bayesian-statistical models are exhaustive and consume less memory. Recently, hyper-dimensional computing, an emerging alternative to ML, has been proposed to address the circuit reliability issues [326]. The experiments to estimate transistor electrical characteristics and manufacturing variability on industrial 14nm FinFET Intel instruments demonstrate 4x smaller error with 20x fewer training samples. Thus, ML and more advanced models will play a significant role in reliability estimation in the future.

### 9.3. Yield Estimation and Management

Many complex and interrelated components during the manufacturing process affect the yield of an IC. Yield learning and optimization are critical for advanced IC design and manufacturing. A yield prediction model is necessary to precisely evaluate the productivity of new wafer maps because the yield is directly related to the productivity and the design of the wafer map affects the yield [327]. Many statistical approaches [328, 329] for yield modeling and optimization have been proposed since the 1980s; however, with the uncertainty in nanoscale fabrication and the growing complexity of the process, large volumes of data are being generated daily, traditional approaches have limits in extracting the full benefits of the data. Even the most complex sophisticated process results in poor exploitation of data. AI/ML could aid in continuous quality improvement in a large and complex process.

Cycle time(CT) is one of the critical performance measures of the semiconductor production line. It is a mandate to understand the key factors influencing CT for its effective reduction and yield enhancement. A data-driven approach to predict the CT understanding the key factors influencing it is proposed in [330, 331]. The approach of data mining and ML can be used for analyzing the extracted information and knowledge from different stages of manufacturing for troubleshooting and defect diagnosis, which decreases the turnaround time. Learning approaches are proposed in [332] for yield enhancement. A backend final test yield prediction at the wafer fabrication stage using a Gaussian Mixture Models (GMM) clustering approach through a weighted ensembled regressor is proposed in [333]. Yield prediction at an early stage helps in cost reduction and quality control. However, there are some limitations to GMM - sensitive to initial guesses of parameters and high chances of getting stuck at the local minimum. As an extension to this, [334] propose a final test yield optimization approach through wafer acceptance test parameters' inverse design.

Classification aids in minimizing wafer yield loss and package yield improvement by thoroughly analyzing data across fab measurements, wafer tests, and package tests [335]. In [336], an ROI-based (return on investment) wafer productivity model using DNNs as a yield prediction technique and differential evolution for optimization is proposed. The DNNs are trained using geometric features of dies. A DNN approach exploits spatial relationships among positions of dies on a wafer and die-level yield variations collected from a wafer test to predict the yield for pre-evaluating the productivity of new wafers [327].

ML is gradually being utilized in yield prediction and optimization and is still in its early stages. There is scope for significant growth in using various ML techniques in yield enhancement.

## 10. AI at Testing

VLSI testing is the process of detecting possible faults in an IC after chip fabrication. It is the most critical step in the VLSI design flow. The earlier a defect is detected, the lesser the final product cost. Rule of 10 [337] states that the cost of fault detection increases by order of 10 moving from one stage to the next in the IC design flow. Improving the yield is a necessity for any company; shipping defective parts can destroy a company's reputation [338]. Almost 70% of the design development time and resources are spent on VLSI testing. Different stages of the design flow involve different testing procedures. Broadly different levels of testing are functional verification testing, acceptance testing, manufacturing testing, wafer level testing, packaging level testing, and so on [339]. We highlight the significant areas of testing that has AI/ML contributions.

### 10.1. Functional Verification

Functional verification is verifying that a design conforms to its specifications. A set of input vectors is provided to the CUT (circuit under test), and its output is compared to the golden output of the specification for checking the possibility of faults. Functional verification [340] is very difficult because of the sheer volume of possible test cases, even in a simple design. Manufacturers generally employ a random test pattern generator [341], which provides significant fault coverage; however, it may only cover some of the faults and has a very long runtime. ML is being used to predict the best test set to achieve the maximum fault coverage with a minimum number of test vectors. In [342], the nearest neighbor algorithm was used to generate efficient patterns for BIST (built-in self-test) test pattern generation, improving the fault coverage. This algorithm detects random pattern-resistant faults and produces test patterns directed toward them. Bayesian networks were used [343, 344] to predict the test pattern set. A Bayesian network is a graphical representation of the joint probability distribution for a set of variables. The Bayesian network model describes the relations between the test directives and coverage space and is used to achieve the required test patterns for a given coverage area. It was further enhanced by clustering the coverage events and working on them as a group. Hughes et al. [345] proposed an ML approach for functional verification, where a NN model is used with RL to track the coverage results of a simulation and, after that, to generate a set of verification input data recommendations, which will increase the probability of hitting functional coverage statements and identifying hard-to-hit faults while adjusting itself.

The initial stage of the CUT greatly impacts the time taken and the ability of stimuli generators to generate the requested stimuli successfully. Some initial states can lead to poor fault coverage, resulting in faulty products. Bayesian networks are employed to automatically and approximately identify the region of favorable initial states; however, they require a certain level of human guidance to select one of the initial states [346]. Identification of power-risky test patterns is also essential, as excessive test power can lead to failure due to IR drop and noise. However, simulation of all the patterns is impossible due to the long runtime. Thus, the pre-selection and creation of a subset of patterns are crucial. Dhotre et al.[347] proposed a transient power activity metric to identify potentially power-risky patterns. The method uses the layout and power information to rank the patterns approximately according to their power dissipation and subsequently uses a K-means clustering to cluster all the instances with concentrated high switching activity.

The application of ML can be extended to delay test measurements as well. Wang et al. [348, 349] proposed models for $F_{max}$ prediction based on the results of structural delay test measurements to determine the optimum conditions for improving the correlation between the golden reference and potential low-cost alternative for measuring the performance variability of the chip design. The performance and robustness of the proposed methodology with a new dataset pruning method, called "conformity check," is demonstrated on a high-performance microprocessor design using KNN, least-squares fit, ridge regression, SVR, and Gaussian process regression (GPR) models. GPR has proven effective in achieving accurate functional and system $F_{max}$ prediction.

In [350], an explainable ML approach called Sentences in Feature Subsets (SiFS) for test point insertion (TPI) is proposed. The proposed ML methodology can also apply to human-readable classification in EDA. An ANN-guided ATPG (automatic test pattern generation) [351, 352] proposed in the recent past reduces the backtracks for PODEM and improves the backtraces, particularly in reconvergent fault-free circuits with reduced CPU time. The training parameters include input-output distances and testability values from cop (controllability and observability program) for signal nodes. Unifying ANN for ATPG incurs a one-time cost, after which ML imparted to ATPG can have long-term benefits. Design2Vec, a deep architecture that learns representations of RTL syntax and semantic abstractions of hardware designs using a GNN, is proposed in [353]. These representations are applied to several tasks within verification, such as test point coverage and new test generation to cover desired points. Pattern identification and reordering method are presented in [354]. An ML algorithm was used to select the most effective test patterns, and then an optimal pattern sequence was determined using the weighted SVMRANK (SVM Rank classification) algorithm. Experiments show time-saving of 3.89 times at the expense of 2% prediction accuracy. In [355], a KNN method is proposed to divide the test patterns into valid and invalid patterns and then use only valid patterns to reduce the test time. Experiments show that compared to the traditional method; this methodology reduces the test time by 1.75 times. Chen et al. proposed an RL-based test program generation technique for transition delay fault (TDF) detection [356].

Even though ML has shown significant progress and promise for functional verification, more is needed to perfect accuracy and human intervention. Intelligent data collection procedures and a novel feature extraction scheme with MI should be inducted into CAD tools as initial steps for IC testing to become fully automated.

### 10.2. Fault Diagnosis

After functional verification, the following test procedure identifies the fault location and type, called fault diagnosis. Traditionally, this step is not fully automated, and the engineer's experience and intuition play a part in developing the test strategies. At present, digital circuits and systems are almost fully automated and have been extensively explored. In contrast, analog circuits are more difficult to diagnose; over the last few years, extensive research on analog fault diagnosis has been conducted, and many ML models have been reported. Most of these models focus on obtaining the output response of a circuit for the test pattern; different pre-processing techniques are applied, bore applying this data as input to the ML model, which attempts to classify the fault. In [357], the Fourier harmonic components of the CUT response are simulated from a sinusoidal input signal and supplied to a two-layer MLP, which attempts to identify the fault. Additionally, a selection criterion for determining the best components that describe the circuit behavior under fault-free (nominal) and fault situations is used and provided as input to a NN [358]. The NN, along with clustering, classifies the faults into a fault dictionary.

In [359], wavelet transform along with PCA was used as a pre-processing technique to extract the optimal number of features from the circuit node voltages. A two-layer NN was trained on these features to the probability of input features belonging to different fault classes. This model yields a 95%-98% accuracy on nonlinear circuits. It was improved in [360] by dividing the circuit successively into modules. At each stage of module subdivision, a NN is trained to determine the sub-module that inherits the fault of interest from the parent module. It led to an increase in the training efficiency of the NN, resulting in 100% accuracy in the classification. A novel anomaly detection technique [361] for post-silicon bug diagnosis was proposed to detect aberrant behaviors or anomalies and identify a bug's time and location. This algorithm comprises two stages. In the first stage, it collects data online by running multiple executions of the same test on the post-silicon platform, after which it records compact measurements of the signal activity. In the second stage, it analyzes the data offline. The authors measured the amount of time the signal's value was one during the time step and applied ML to the measurements to locate the bug. The fundamental goal of testing is to determine the defects' root causes and eliminate them.

Automatic defect classification, which has existed for several years, has been revolutionized by ML in terms of speed and accuracy, although ML-based defect analysis is still not ideal for industrial standards. Much focus is needed on automated defect analysis to locate the root cause of the defect.

### 10.3. Scan Chain Diagnosis

Scan chain structures are widely used in VLSI circuits under design for testing. They increase the fault coverage and diagnosability by enhancing the controllability and observability of the digital circuit logic [362]. Figure 13 shows the design of a preliminary scan chain. During normal circuit operation, these structures function like a regular flip-flop, and during testing, they shift and capture data at intermediate nodes, aiding the identification of the fault location. However, the circuit cannot be tested if a fault occurs in the scan chain. Therefore, scan chain diagnosis is very crucial. Traditionally, many special-tester-based and hardware-based diagnostic techniques were used. Although they provide high accuracy, they are computationally expensive and time-consuming. Recently, software-based diagnostic methods have attracted significant attention; however, these methods do not provide satisfactory results. ML is widely used in scan chain diagnosis to achieve sufficient resolution and accuracy.

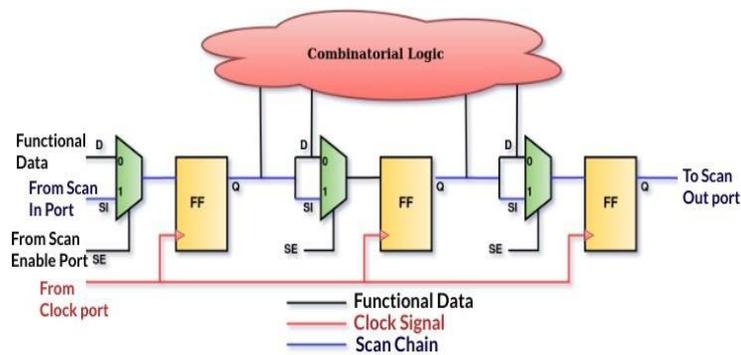

**Figure 13:** A Typical Scan Chain Design

An unsupervised-learning model was proposed [363], where a Bayesian model was employed for diagnosis. The failing probabilities of each scan cell were supplied as input to the model, which partitioned the scan cells into multiple clusters. After that, the defective scan cell is found at the boundaries of adjacent clusters. This model yielded 100% accuracy for both permanent and intermittent faults, although only for single stuck-at faults. ANNs have come into use recently, providing sufficient resolution and accuracy. For example, in [364], a coarse global neural network was used to select several suspected scan cells (affine group) from all the scan-chain cells, and a refined local neural network to identify the final suspected scan cell in the affine group. This successively refined focus increased the resolution and accuracy but significantly increased the training time due to multiple networks. A two-stage NN model was proposed to identify the exact location of a stuck-at-fault and transition fault in [365]. The 1st stage ANN trained with entire scan data with all faults predicts a scan window with successive candidates. The 2nd stage ANN analyzes the fail data locally to identify the exact fault location.

Liu et al. proposed RF classification to predict test chip design exploration synthesis outcomes [366]. In [367], a DT-based screening method is proposed to predict unreliable dies that would fail the HTOL (high-temperature operating life) test [367]. The HTOL test is a popular test to determine the device's intrinsic reliability and predict the device's long-term failure rate and lifetime of the device [368]. SVM and autoencoder-based early stage system level testing (SLT) failure estimation reduces the testing cost by 40% with a minor impact on defective parts per million (DPPM) [369]. In addition, adaptive test methods that analyze the failing data and test logs, dynamically reorder the test patterns and adjust the testing process bring down the testing cost by several orders [370, 371].

The state-of-the-art DL for IC test (GCNs (Graph Convolutional Networks) and ANNs in particular) is discussed in [372]. The work systematically investigates the robustness of ML metrics and models in the context of IC testing and highlights the opportunities and challenges in adopting them. A novel physics-informed neural network (PINN) to model electrostatic problems for VLSI modeling applications achieves an error rate of 9.3% in electric potential estimation without labeled data and yields 5.7% error with the assistance of a limited number of coarse labeling data [373]. The paper also highlights the implementation of ML models for data exploration for IC testing and reliability analysis. In a survey of ML applications on analog and digital IC testing, significant challenges and opportunities are presented [374].

We observe that deep NNs, GNNs in particular and Bayesian networks are the most suitable approaches to act as an alternative to various laborious manual testing procedures.

## 11. Sources of Training data for AI/ML-VLSI

The techniques of AI/ML would aid in solving many challenges in the IC industry. Nevertheless, the limited data availability for training the necessary algorithms is a known difficulty in VLSI domain. Although there is a plethora of tools for designing, manufacturing, and testing VLSI circuits, a systematical way of capturing relevant and sufficient data for training AI/ML algorithms still needs to be solved. A structured methodology for automated data capture across different design levels needs to be incorporated into the IC design flow to resolve the challenge of data scarcity to a certain extent.

This section presents a brief on sources of training data explored and implemented in literature for future research interest (Fig. 14). SIS is an interactive tool for synthesizing and optimizing sequential circuits that produces an optimized netlist in the target technology [375]. Benchmark circuits to analyze hardware security are available at Trust-HUB [376]. The research community utilized EDA tools from Cadence, Synopsys, and Mentor Graphics, while ISCAS and ISPD benchmarks were used by many to generate training datasets and for testing/model validation.

## 12. Challenges and Opportunities for AI/ML in VLSI

The dimensions of devices are decreasing; however, as we approach atomic dimensions, many aspects of their performance deteriorate, e.g., leakage increases (particularly in the sub-threshold region), gain decreases, and sensitivity to fluctuations in manufacturing processes increase drastically [7]. This results in less reliability and yield. The growing process variability and environmental sources of variation in nanometer technology are leading to the deterioration of the overall circuit performance. Modeling these effects based on worst-case process corners would no longer be valid as most parameters vary statistically. Moreover, many parameters exhibit complex correlations and wide variances.

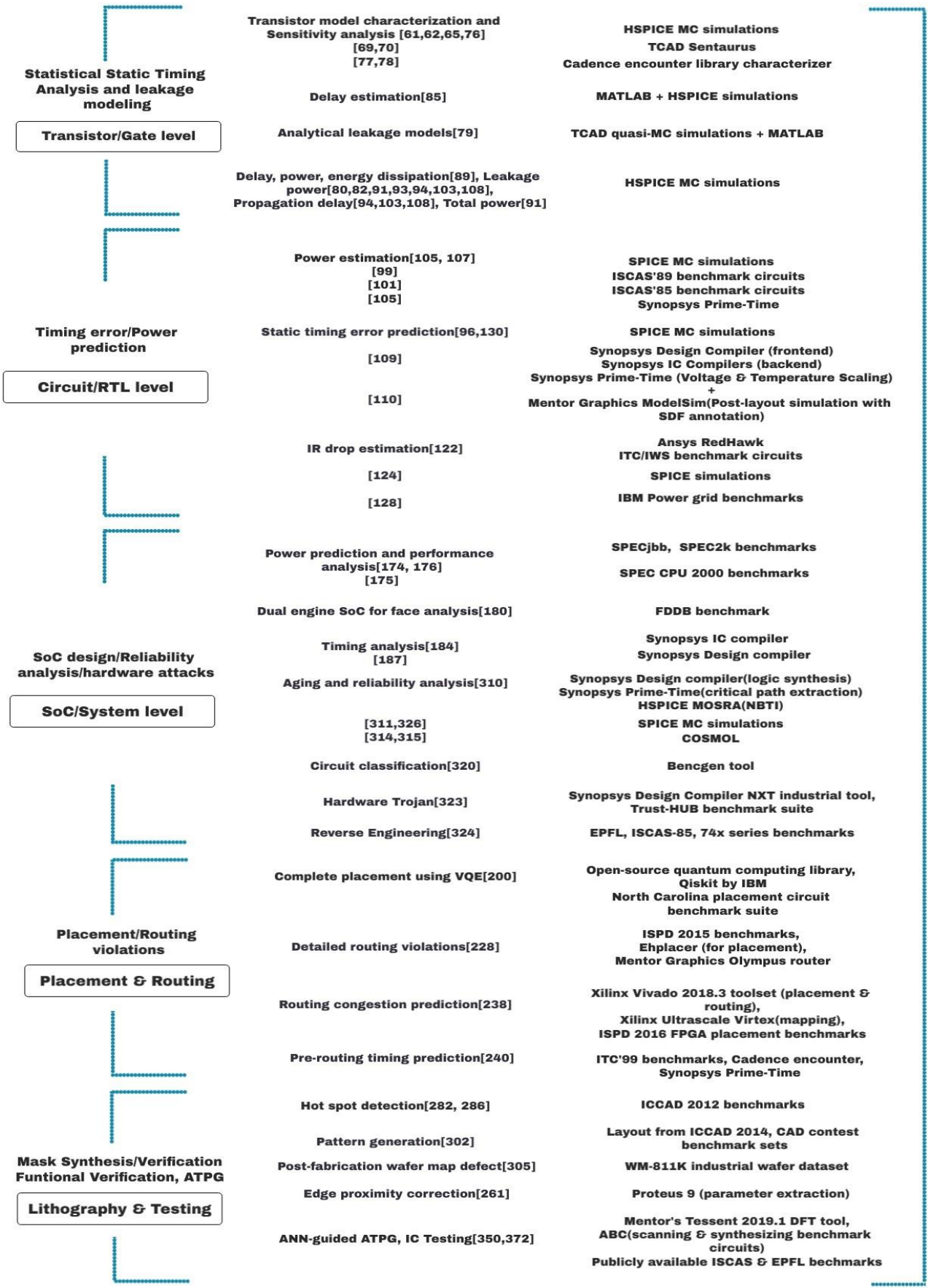

**Figure 14:** Sources of training data in the literature

Presently, the computationally efficient methods for estimating the outputs corresponding to the inputs are some of the areas attracting significant interest in the field of circuit modeling of VLSI–CAD. To maximize chip relia- bility and yield, each design in VLSI is optimally tuned to consume and dissipate low power, occupy a minimum area, and achieve high throughput. Device models coupled with circuit simulation tools significantly improve design productivity, providing insights for improving the design choices and circuit performance [7]. Accurate and fast estimation techniques are required during circuit design and modeling to estimate and verify the effect of the process variations on the circuit output; this can aid the incorporation of corrective measures/methodologies to improve the yield, thereby guaranteeing the design quality. The primary challenge under process variations is to identify the dominant parameters causing the variations, estimate the relationship between the dominant parameters and the circuit performance parameters, develop models for performance evaluation, and incorporate these models into design tools. This problem is more pronounced in the nanometer regime with the increased complexity of digital design. Traditional models estimating the circuit performance comprise many parameters and have complicated equations that significantly slow the simulation speed. At the current technology nodes, there is a need for compact device models with essential capabilities of scalability and universality (i.e., the ability to support different technologies). Nevertheless, one can see many opportunities to address these challenges.

Surrogate ML/AI models provide solutions to these problems. These models forecast the device performance and can be easily extended to circuit-level and system-level design and analysis. Such models have been proposed in the literature to improve the turnaround time, and yield of ICs [91, 93]. This learning methodology can also be applied for post-layout simulation and ECO, aiding the achievement of timing closure [120]. Surrogate AI/ML models offer comparable simulation rates to traditional EDA tools with reasonable accuracy. Potential risks in the advanced silicon nodes can be estimated and analyzed with prior design data using ML algorithms. These algorithms can better capture complex electrical behavior in advanced technology nodes than traditional EDA tools. The best methodologies for incorporating computationally effective AI/ML models into VLSI–CAD design tools need to be explored.

Estimation and analysis of the subsystem behavior are also crucial in IC technology. For instance, accurately estimating the subsystem power consumption of commercial smartphones is necessary for various applications in many research areas. In this regard, learning algorithms will improve the end-to-end performance, promoting a high utilization ratio and high data bandwidth [377]. Memory designs on nanometer technology nodes are becoming increasingly challenging because they are the smallest devices on the chip and are thus affected the most in terms of functionality and yield. Increasing the inter-die and intra-die variabilities will exacerbate the cell-stability concerns. The AI/ML approach also increases the statistical analysis rate of memory designs. The learning strategies of AI/ML have been extended to high-level SoC designs [378] in the past. Kong and Bohr [379, 380] discuss the survey of design challenges in the nanometer regime. A vast network of AI is employed in hardware acceleration to implement dynamic high-level digital circuits onto the hardware. High-speed VLSI hardware systems provide the necessary driving capability for AI/ML algorithms to achieve their maximum potential. Dense NNs used extensively in embedded systems, such as IoT sensors, cars, and cameras, need high classification speeds, which are possible with high-speed hardware accelerators [381, 382]. In-memory-based computing by IBM [383] demonstrates how the completion speed of tasks by ML can be increased significantly with reduced power consumption. The realization of AI/ML learning algorithms in hardware reduces the learning time and increases the speed of the prediction process by many orders of magnitude. Interconnect datasets need to be built carefully with many SoC parameters, floor planning, routing constraints, and clock characteristics, creating ample design space for exploration. Current GPUs offer high acceleration rates with parallel computing facilities and superior performance for large design spaces.

Present ASIC design methodologies break down in light of the new economic and technological realities; new design methodologies are required for which the physical implementation of the design is more predictable. A database and interface, from design-to-manufacturing to effectively managing the parameter variability and increasing the data volume, must be provided [379]. A paradigm shift in CAD tool research is required to manage complex functional and physical variabilities. The AI/ML algorithms can be unfolded to the CAD-tool methodologies in physical design to manage the involved complexities. Data mining approaches, such as clustering and classification, can be imbibed into VLSI partitioning, paving a new route for recognizing hidden patterns in data and predicting the relationships between the attributes that enable forecasting outcomes [384].

Similarly, learning strategies can be applied to find cost-effective solutions for placement and routing [23]. Reducing the design cost of an IC is the primary driving force for downscaling [385]. The continued shrinkage of logic devices has brought about new challenges in chip manufacturing. It is increasingly difficult to resolve fine patterns and place them accurately on the die, particularly at sizes below 20 nm. ML techniques can be utilized in the chip-manufacturing-process-optimized compact-patterning models in the lithography process, mask synthesis, and correction and can be extended to physical verification to validate the design's manufacturability. For automated recovery and repair and big data debugging, the challenges in chip manufacturing need to be addressed. Post-silicon validation is also possible using ML algorithms with available training data from the pre-silicon stage. The cost of testing a VLSI chip/subsystem can be reduced using AI algorithms. For instance, finding an efficient solution for rearranging the test cases using AI heuristic search algorithms can reduce power dissipation during testing [386].

Having stated that there are many critical problems, such as high variability and deteriorated reliability, a wide variety of AI and ML approaches - supervised/unsupervised/semi-supervised learning; NNs, MLP structures [387], [43], [59] - CNNs and deep learning [388], provide opportunities to solve the numerous problems and challenges in the field of VLSI design. There is a trade-off between selecting suitable algorithms and architectures with available training data and other model constraints. Fitting these new techniques in the classical flow of VLSI is another big challenge. Another issue is the availability of standardized, licensed ML algorithms with a thorough debugging facility. High-yielding implementations are achievable by critically channeling ML designers' domain knowledge with CAD designers.

The availability of limited training data can be maximally solved if the data flow across the design cycle can be effectively captured and explored. The chip designing industries should understand the importance of systematic generation, the capture of data, the incorporation of distributed bid data systems for chip workflows, and data-driven optimizations to accelerate the quality, cost, and time of results [389]. It could be beneficial to have benchmark datasets for AI/ML training for future research and automated IC design flow development. To address the dearth of training data, the critical challenge for employing AI/ML, standardized statistical training data for circuit modeling should be developed. Such open-source contributions in the VLSI community help to address the challenges more effectively. Further, researchers can use them as a baseline and rapidly progress [390].

Different abstraction levels in the design flow, ranging from circuit design to chip fabrication and testing, inherently comprise numerous models relating inputs to outputs. An enormous amount of data flows across billions of devices or components integrated/to be integrated on the chip [23]. The complex I/O relationships between the components, processes and various abstraction levels within each abstraction level can be explored via AI/ML algorithms using the information accumulated during different kinds of simulations/analyses. Further, we need to analyze the data streams associated with file operations, which clustering algorithms can use to deliver high application performance. AI/ML solutions can be employed in VLSI–CAD for design-flow optimization.

Future advancements in differential programming and quantum ML approaches can lead to incredible breakthroughs in the EDA industry.

AI/ML in VLSI

AI/ML in VLSI

D Amuru et al.: *Preprint submitted to Elsevier* Page 34 of 41

AI/ML in VLSI